\documentclass[journal]{IEEEtran}

\usepackage[pdftex]{graphicx}
\usepackage{amsmath}
\usepackage{algorithmic}
\usepackage{algorithm}
\usepackage{listings}
\usepackage{hyperref}
\hypersetup{colorlinks=true,urlcolor=black,linkcolor=black,citecolor=black}
\usepackage{dirtytalk}
\usepackage{cleveref}
\usepackage[utf8]{inputenc}
\usepackage{subfigure}
\usepackage{microtype}
\usepackage{enumitem}
\usepackage{amssymb}

\usepackage{color}
\usepackage{xcolor}
\usepackage{booktabs}
\usepackage{array}
\usepackage{rotating}
\usepackage{adjustbox}
\usepackage{array}

\newcolumntype{R}[2]{%
    >{\adjustbox{angle=#1,lap=\width-(#2)}\bgroup}%
    l%
    <{\egroup}%
}
\usepackage{flushend}
\usepackage[colorinlistoftodos,prependcaption,textsize=tiny]{todonotes}
\newboolean{showcomments}
\setboolean{showcomments}{true} 
\ifthenelse{\boolean{showcomments}}
{\newcommand{\nb}[2]{
		\fcolorbox{black}{yellow}{\bfseries\sffamily\scriptsize#1}
		{\sf\small$\blacktriangleright$\textit{#2}$\blacktriangleleft$}
	}
	
}
{\newcommand{\nb}[2]{}
	
}

\newcommand\parhead[1]{\vspace{.26mm}\noindent\textbf{{#1}.}}

\newcommand{\secref}[1]{Sec.\,\ref{#1}}
\newcommand{\Figref}[1]{Figure\,\ref{#1}}
\newcommand{\figref}[1]{Fig.\,\ref{#1}}
\newcommand{\tabref}[1]{Table\,\ref{#1}}

\newcommand{\xcite}[1]{}

\usepackage[normalem]{ulem} 

\begin{document}

\title{A Driver-Vehicle Model \\for ADS Scenario-based Testing}

\author{Rodrigo~Queiroz,~Divit~Sharma,~Ricardo~Caldas,~Krzysztof~Czarnecki,~Sergio~Garcia,~Thorsten~Berger, and~Patrizio~Pelliccione%

\thanks{R. Queiroz, D. Sharma, and K. Czarnecki are with the University of Waterloo, Canada. R. Caldas, S. Garcia, and T. Berger (secondary affiliation) are with Chalmers\,$|$\,University of Gothenburg, Sweden. T. Berger (primary affiliation) is with Ruhr University Bochum, Germany, P. Pelliccione is with Gran Sasso Science Institute (GSSI), Italy}%
}

\markboth{IEEE Transactions on Intelligent Transportation Systems}
%
{Shell \MakeLowercase{\textit{et al.}}: Bare Demo of IEEEtran.cls for IEEE Journals}
%



\maketitle

\begin{abstract}
Scenario-based testing for automated driving systems (ADS) must be able to simulate traffic scenarios that rely on interactions with other vehicles. 
Although many languages for high-level scenario modelling have been proposed, they lack the features to precisely and reliably control the required micro-simulation, while also supporting behavior reuse and test reproducibility for a wide range of interactive scenarios. To  fill  this  gap between  scenario  design and execution, we propose the Simulated Driver-Vehicle (SDV) model to represent and simulate vehicles as dynamic entities with their behavior being constrained by scenario design and goals set by testers. The model combines driver and vehicle as a single entity. It is based on human-like driving and the mechanical limitations of real vehicles for realistic simulation. The model leverages behavior trees to express high-level behaviors in terms of lower-level maneuvers, affording multiple driving styles and reuse. Furthermore, optimization-based maneuver planners guide the simulated vehicles towards the desired behavior. Our extensive evaluation shows the model’s design effectiveness using NHTSA pre-crash scenarios, its motion realism in comparison to naturalistic urban traffic, and its scalability with traffic density. Finally, we show the applicability of our SDV model to test a real ADS and to identify crash scenarios, which are impractical to represent using predefined vehicle trajectories. The SDV model instances can be injected into existing simulation environments via co-simulation.
\end{abstract}


%
\IEEEpeerreviewmaketitle

\section{Introduction}
\noindent
\looseness=-1
Testing automated driving systems (ADS) requires simulating a wide range of operating scenarios to ensure an ADS's safety and conformity to traffic regulations and industry standards. As the responsibility for the driving task shifts from the human driver to the ADS\,\cite{sae:2014:j3016}, the system is required to handle interactions with other road users, especially human-operated vehicles. 
Test scenarios must reflect how these dynamic interactions between the subject system (a.k.a. \emph{ego} vehicle) and other vehicles can unfold in real traffic.

\begin{figure}[t]
    
	\centering
	 \includegraphics[width=1.0\linewidth]{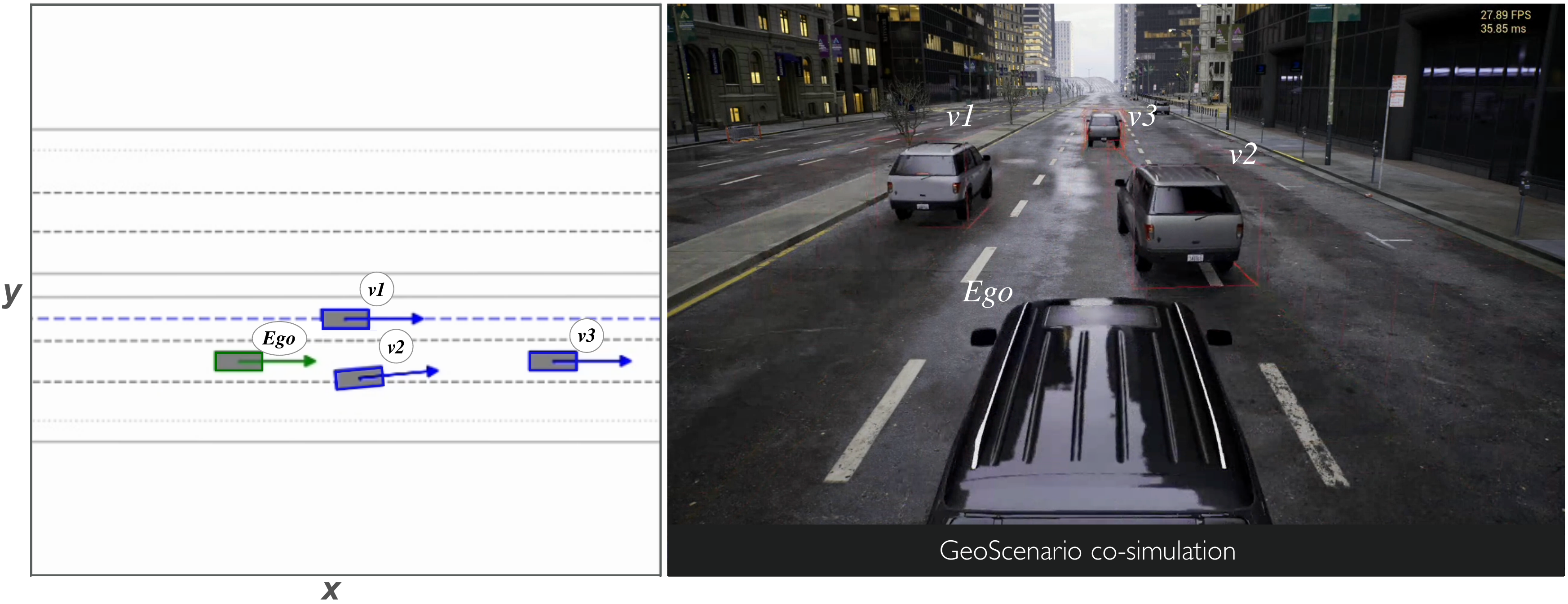}
	 \caption{A challenging interaction between ego and human-operated vehicles based on a pre-crash scenario from NHTSA\,\cite{nhtsa:2007:crashtypology} and using our SDV model in simulation: (left) in the map frame, the SDV Model as $v_{2}$ performs the cut-in maneuver targeting ego, and (right) a high-fidelity co-simulator renders the scene.}
	 \vspace{-4 mm}
	\label{fig:cutinscenario}
\end{figure}

\looseness=-1
\Figref{fig:cutinscenario} shows a near-collision of ego with $v_{2}$ cutting-in before, taken from the National Highway Traffic Safety Administration's (NHTSA) pre-crash scenario catalog\,\cite{nhtsa:2007:crashtypology}. The cut-in maneuver of $v_{2}$ triggers reactions by other close vehicles, with ego's reaction strongly influencing how the scenario unfolds. Testing collision avoidance in such scenarios requires models able to represent and simulate traffic dynamics, including the interactions between ego and other human-operated vehicles.

\looseness=-1
Many domain-specific languages (DSLs) \cite{dslbook,lammel2018software} for scenario-based testing have emerged.
These DSLs include models for representing test scenarios. Testers design such scenarios by defining behaviors of human-operated vehicles, and then executing them in simulation tools. However, these DSLs are often limited to relatively simple models, for instance, replay of pre-recorded trajectories\,\cite{queiroz:19:geoscenario}, event-based orchestration to directly manipulate the vehicle state\,\cite{online:openscenario,online:msdl}, and narrow behavior models (e.g., vehicle following\,\cite{kesting_enhanced_2010}). As a result, testers may have limited control over the precise movement of the simulated vehicles and the resulting behaviors may vary between simulation tools, hurting test reproducibility and the validity of test results.

\looseness=-1
To bridge the gap between scenario design and execution, we contribute the \emph{GeoScenario Simulated Driver-Vehicle (SDV) model} to specify and simulate realistic behavior of human-operated vehicles in ADS scenario testing. SDV offers high expressiveness, execution accuracy, scalability, and reuse. It extends the scenario-definition language GeoScenario\,\cite{queiroz:19:geoscenario} with human-operated vehicles as dynamic agents in both scenario representation and simulation execution. It implements a driver behavior model inspired by Michon et~al.\,\cite{michon:1985:drivermodel}, including (i) route selection as a strategic decision, (ii) maneuver selection as a tactical decision, and (iii) maneuver implementation as an operational decision. Specifically, the maneuver selection logic is expressed using behavior trees\,\cite{colledanchise:2018:book,ghzouli.ea:2023:tse}, offering modularity and reuse. The maneuvers are implemented using an optimization-based trajectory planner, which guides the simulation towards achieving the scenario's test objectives. 
The executed maneuvers can be configured by simulation engineers to reflect different driving styles, subject to the human and physical limitations of actual vehicles.

\looseness=-1
We evaluate our model's (i) scenario design effectiveness, which includes expressiveness, execution accuracy, and reuse, using NHTSA pre-crash scenarios; (ii) motion realism in comparison to naturalistic urban traffic; (iii) scalability with traffic density; and (iv) practical applicability to test an actual ADS. 
The results show that our model can successfully express, achieving levels of model reuse of over 80\,\%, and accurately execute all eighteen NHTSA vehicle-to-vehicle pre-crash scenarios (except one variant), while only four scenarios are effectively expressible using predefined trajectories, which is our baseline. We also show that, after calibration, the model is capable of producing maneuver decisions and trajectories that closely resemble those from recorded real-world traffic. The model also scales in scenarios with up to 10--20 simultaneous and highly interactive vehicles in real-time simulation. Finally, we demonstrate the model's applicability to test an ADS software stack in simulation, which has been tested on public roads, and reveal collision scenarios that cannot be expressed using the baseline.

\looseness=-1
In summary, our paper contributes:
\begin{itemize}
     \item a novel simulation model for human-operated vehicles, that combines behavior trees with an optimization-based trajectory planner to provide a highly-expressive, controllable, realistic, reusable, and scalable scenario representation for ADS testing;
     \item a set of experiments to support our claims about the qualities of the model;
     \item an open-source reference implementation of the model, which can be integrated with any simulation environment in co-simulation mode.\footnote{\url{https://github.com/rodrigoqueiroz/geoscenarioserver}}
\end{itemize}

\section{Background and Related Work}
\label{sec:background}

\parhead{Scenario-Based Testing} 
Kaner et al.\,\cite{kaner:2001} define \textit{scenario-based testing} as the dominant paradigm of black-box testing, where scenarios are used to check how the system copes with both nominal and off-nominal situations. 
In the automotive context, ISO 26262\,\cite{iso:26262} and ISO 21448\,\cite{iso:21448} guide the development of safety-critical electrical/electronic vehicle systems and mandate the use of scenarios in validation activities.

Scenarios are designed based on expert knowledge and on the traffic situations the ADS must be able to cope with, or by reproducing and augmenting situations collected from traffic databases. 
For example, CommonRoad~\cite{althof:2017}, a benchmark for motion planners, provides  scenarios extracted from NGSIM data\,\cite{punzo:2011:ngsim}. 
A scenario can also be systematically generated to achieve specific test goals, e.g., lead the system to trigger a certain behavior, such as an emergency maneuver, or find a critical situation leading to a crash. For example, Abdessalem et al.\, \cite{abdessalem:2016:testmo,abdessalem:2018:testevo} use evolutionary optimization methods combined with surrogate model learning to find crash scenarios.

\parhead{Scenario Representation and Driver Behavior} 
\noindent
Multiple tool-independent DSLs have emerged recently, providing a formal definition of scenario structure, behavior, test conditions, and pass/fail criteria to support scenario-based testing in simulation. The goal is to offer a uniform representation and semantics across methods and tools. The scope and structure of each language vary, but fundamentally they all define how vehicles behave in traffic and orchestrate interactions with ego that must be executed by a simulation tool during the test.

OpenScenario\,\cite{online:openscenario} is a standard managed by the Association for Standardization of Automation and Measuring Systems (ASAM). The format describes dynamic content in driving simulation applications in combination with OpenDRIVE\,\cite{online:opendrive}, which specifies the road structure. It covers traffic and driver behavior, weather, environmental events, and other features. 
It includes the description of a driver, but there is no model for driver behavior in any form other than ``road following." The standard also does not contain maneuver models or a vehicle model. Maneuvers are described in terms of \textit{actions} (e.g., change the vehicle's position or speed), and trajectories (defined as a polyline, clothoid, or spline). 

\looseness=-1
The Measurable Scenario Description Language (MSDL)\,\cite{online:msdl} expands the concepts of OpenScenario. The language uses \textit{modifiers} to change the behavior of the agents similarly to \textit{actions} from OpenScenario. It introduces parameter variability (a range instead of a single value) along with constraints to narrow down values and connect multiple parameters (e.g., velocity of vehicle A is between 10 and 20\,m/s and less than vehicle B). The language supports generating concrete scenarios by picking random values while obeying the constraints.

Other formats are Scenic\,\cite{fremont:2019:scenic}, Scenario Description Language (SDL)\,\cite{zhang:2020:sdl}, and  SceML\,\cite{2020:sceml}. A common trait amongst them is that they are primarily declarative languages. They define ``what'' must happen in a scenario during key events without specifying ``how.'' Their approach relies on external simulation models to handle the execution.

\looseness=-1
Finally, GeoScenario\,\cite{queiroz:19:geoscenario} provides mechanisms to represent road users and an orchestration system to allow testers' control of how they interact with ego. The language tackles the multi-agent orchestration via triggers, but is limited at the individual vehicle behavior to select among predefined
trajectories specific to the road. Our SDV model extends GeoScenario with interactive and flexible driver behavior.

\parhead{Models for Traffic Simulation}
\noindent
Macroscopic traffic models describe vehicle motion and interaction in terms of flow and density. They are mainly used for large scale simulation over a road network\,\cite{sewall:2010:conttrafficsim}. They are not suitable for street-level vehicle motion and interactions and thus ADS testing. 

\looseness=-1
In contrast, microscopic traffic models can generate vehicle motion and interactions at the individual vehicle level at the cost of limited scalability \cite{chao:2020:trafficsimsurvey}. They are able to encode simple rules that allow a vehicle to follow waypoints or the structure of the road,  avoid frontal collisions by alternating between driving and stopping, and perform maneuvers triggered by conditions \cite{gipps:1998:followmodel, kesting:1999:lcmodel, dosovitskiy:2017:carla}. However, while capturing this reactive behavior, they usually lack enough detail to simulate complex interactions between the vehicle under test and other road users in realistic conditions. For example, they often use simplistic motion limited to a constant velocity throughout a maneuver and disregard the physical limitation of a real vehicle. They also cannot represent complex interactions, such as vehicles responding to merge attempts, using the available road space to navigate around obstacles, or skillfully navigating an intersection with multiple influencing factors (e.g., vehicles, pedestrians, and traffic regulation).

\looseness=-1
Established microscopic models target a particular maneuver, for example, vehicle following\,\cite{gazis_nonlinear_1961,gipps:1998:followmodel,kesting_enhanced_2010,milanes_modeling_2014}, decisions to perform lane changes\,\cite{kesting:1999:lcmodel,gipps_model_1986}, and the execution of lane changes\,\cite{moridpour:2010:lc}. While these models capture details of speed regulation during vehicle following or the parameters of deciding lane changes, they are suitable for testing specific functions and subsystems (for instance, testing adaptive cruise control) in a very constrained environment. They do not cover the complexity of the full driving task required for scenarios in system-level testing of an ADS, including complex decision making among multiple maneuvers and trajectory generation. They can be used to inform the design and parameter setting of the behavior trees in our model, however. 

\looseness=-1
A different approach is to learn models directly from data. For example, a trajectory prediction model trained on recorded traffic data can be run in closed loop as a simulator\,\cite{suo2021trafficsim,bergamini:2021:simnet,igl:2022:symphony,zhong:2022:conddifusion, suo:2023:mixsim, Jiang:2023:CVPR}. Recent approaches allow a degree of controllability of the road users during simulation, e.g., by using conditional models\,\cite{suo:2023:mixsim} or diffusion models with cost functions\,\cite{zhong:2022:conddifusion,Jiang:2023:CVPR} to guide trajectory sampling during inference. While helping to automate scenario creation, the main limitation of purely data-driven approaches is the inherent bias in the data used to build the models.
In particular, driver mistakes and safety-critical scenarios are rare in traffic and thus usually absent from or rare in existing datasets. In fact, programmable behavior models provide an opportunity to generate such rare scenarios and use them to augment the training for data-driven approaches. Finally, while the models can capture the diversity of driving styles in road environments they were trained on, they are difficult to generalize to other environments\,\cite{gilles:2022:uncertainty}.

Thus, scenario-based testing requires executable models that offer high expressiveness, controllability, and realistic behavior---a combination that existing work currently lacks.

\parhead{Behavior Trees}
\noindent\looseness=-1
Behavior trees is a discrete control architecture, which aims to address the shortcomings of finite state machines and their variations, and provide improved modularity, reusability, scalability, and readability\,\cite{colledanchise:2018:book,ghzouli2020sle,ghzouli.ea:2023:tse}. These user-oriented qualities motivate their use to express driving behavior, which has been explored in the past. Several works have proposed using behavior trees to make maneuver decisions within an ADS\,\cite {olsson:2016:btav,tadewos:2019:safebt,Mais21}. Perhaps the closest is BTScenario\,\cite{kang:2022:btscenario}, which uses them to control maneuvers of vehicles in simulation testing. However, BTScenario uses behavior trees to issue driving control inputs directly to a longitudinal and lateral controller. The lack of a trajectory planner makes it impossible to plan flexible and realistic trajectories to avoid static and dynamic obstacles. The work also lacks a systematic evaluation of expressiveness, reusability, motion realism, and scalability. In another work, we used behavior trees to control pedestrians in simulation, where behavior trees set motion objectives for pedestrians moving according to the social force model\,\cite{larter:2022:pedestrian}. To the best of our knowledge, we are not aware of other work that (i) combines behvaior trees with an optimization-based planner to provide a highly-expressive, controllable, realistic, reusable, and scalable scenario representation for ADS testing, and (ii) systematically evaluates such an approach.

\section{The SDV Model}
\label{sec:overview}
\begin{figure*}
    \centering
    \includegraphics[width=0.8\textwidth]{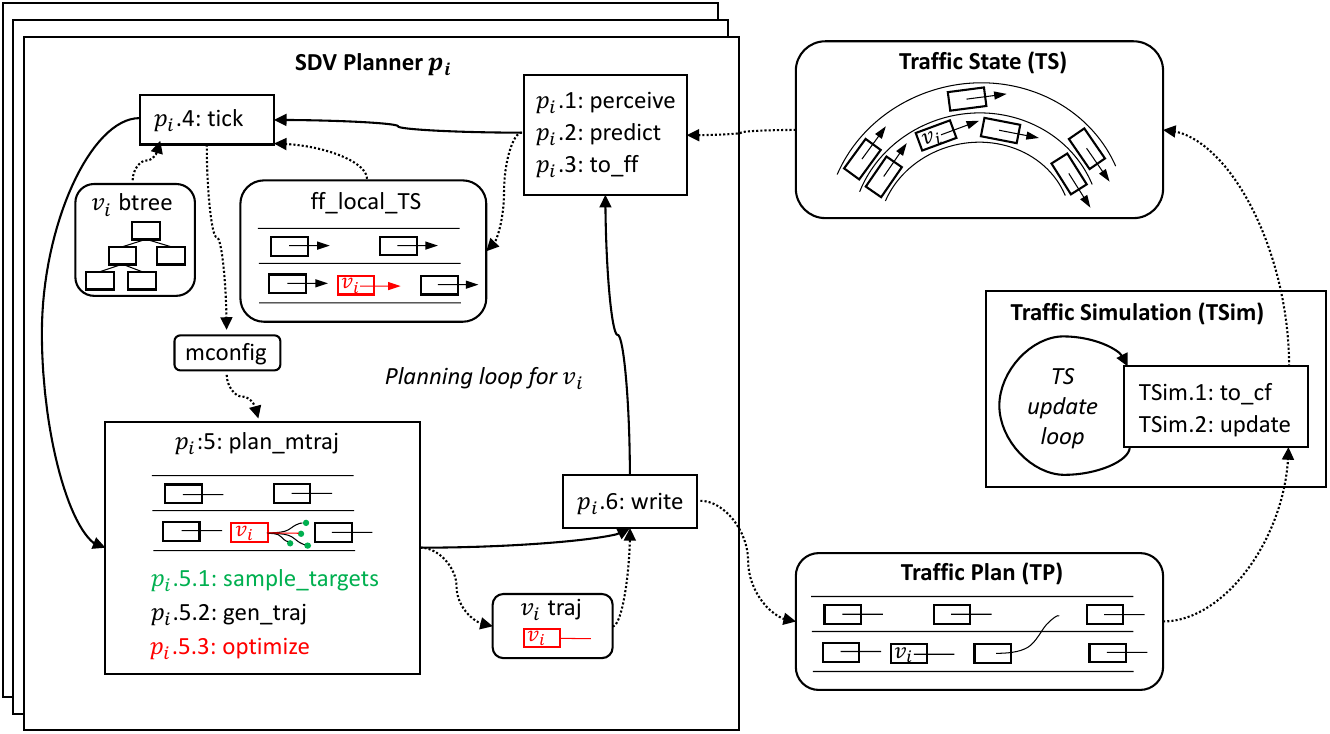}
    \caption{Flow diagram of the simulation. Top-level sharp-cornered rectangles represent processes; nested ones represent procedures. Round-cornered rectangles denote data. Solid arrows represent control flow; dashed ones represent data flow. Arrows attached to vehicles denote velocities; curves denote trajectories.}
    \label{fig:flowchart}
\end{figure*}

\noindent\looseness=-1
We now introduce the concepts and the algorithm of the SDV model (see \figref{fig:flowchart}). The overall simulation consists of (i) a set of simulated road users, each run in a separate process (SDV Planner) that plans its future trajectory, and (ii) a single traffic simulation process (Traffic Simulation) that executes these trajectories. For simplicity and scalability, the model combines driver and vehicle as a single entity (SDV Planner), abstracting away driver inputs, such as steering angle, braking, and throttle.

\begin{figure}[t]
	\centering
	 \includegraphics[width=\linewidth]{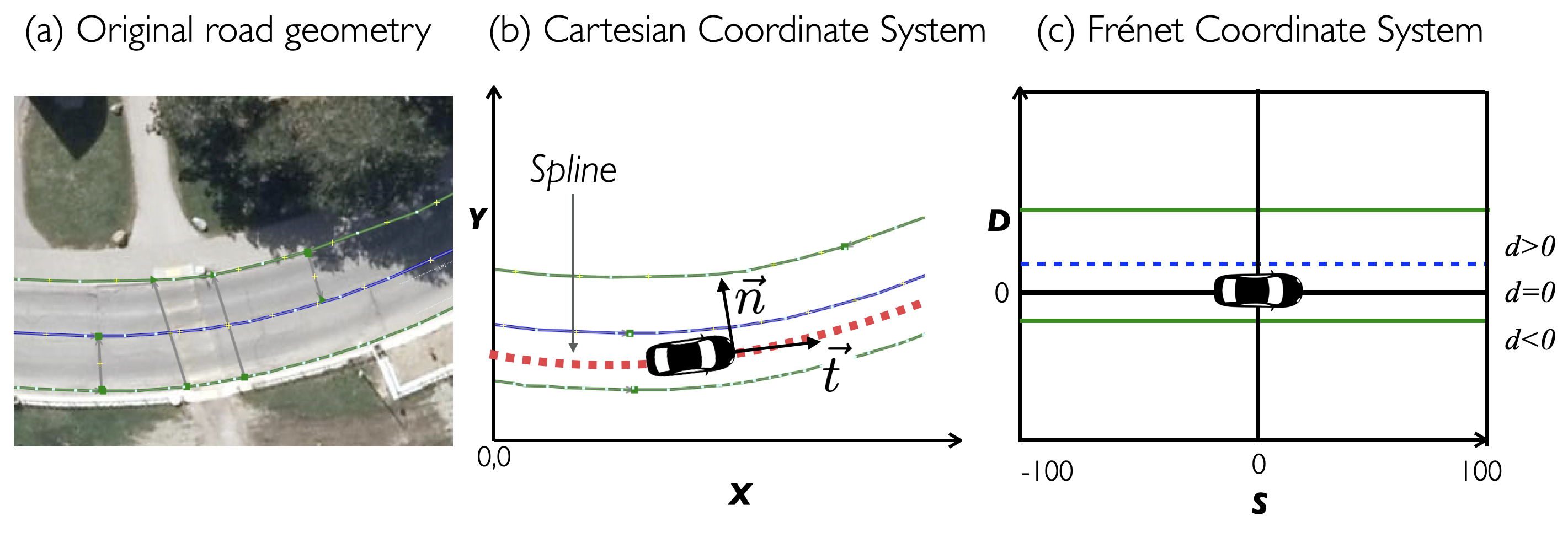}
	\vspace{-8mm}
	\caption{Road Geometry and vehicle displacement from original coordinates are transformed into Frénet Frame using the tangential and normal vectors $\vec{t}$, $\vec{n}$ from the lane centre line (shown in red).}
	\vspace{-2mm}
	\label{fig:frenetframe}
\end{figure}

\looseness=-1
An SDV Planner executes its behavior tree and communicates with the Traffic Simulation using two shared variables: an SDV Planner $p_i$ reads the traffic state (TS) and writes the traffic plan (TP), and the Traffic Simulation reads TP and writes TS. The latter includes the current state of all vehicles, including their coordinates $x,y$ in the global Cartesian frame of the simulation, their first and second time derivatives, and heading $\theta$:
\vspace{-2mm}
\begin{flalign} \label{eq:cartstate}
\textit{VehicleState}_\text{Cartesian}(t)=[x, \dot{x}, \ddot{x}, y, \dot{y}, \ddot{y},  \theta]_t
\end{flalign}
\looseness=-1
The traffic plan includes the future trajectories for all SDVs. Each trajectory is represented in the Frénet reference frame\,\cite{werling:2010:optff} of the respective SDV  (\figref{fig:frenetframe}). This is motivated by the fact that safety requirements on the motion of an on-road vehicle are typically specified relative to its Frénet frame derived from the local lane geometry (see, e.g., Shalev-Shwartz et al.\,\cite{shalevshwartz2018formal}).

\parhead{SDV Planner}
An SDV Planner process is instantiated for each SDV, as indicated by the stacked boxes in \figref{fig:flowchart}. The SDV Planner $p_i$ for vehicle $v_i$ is given a route, represented as a sequence of lane segments that can be legally traversed by the vehicle, and a behavior tree (btree), and it performs a maneuver planning loop with six steps ($p_i$.1--6).
Maneuver planning starts with the procedure $p_i$.1:perceive, which retrieves the current traffic state from the perspective of $v_i$ and simulates perception, including sensor range. The next procedure, $p_i$.2:predict, projects the perceived local traffic state forward to the future simulation time targeted by the current maneuver planning iteration. Prediction uses the previously planned trajectory for $v_i$ but assumes constant velocity for all other vehicles, including externally-simulated ones for which planned trajectories are not observable, such as ego under test. The next step, $p_i$.1:to\_ff, transforms the local traffic state TS into the Frénet frame of $v_i$, resulting in ff\_local\_TS. The frame is defined w.r.t. the center line of the lane (red in \figref{fig:frenetframe}(b)) that $v_i$ is traveling on as part of its route (\figref{fig:frenetframe}(a)). Its origin is the point along this line that is closest to the vehicle. The resulting frame's $S$ axis represents the longitudinal displacement along this center line, and the $D$ axis represents the lateral displacement.

\begin{figure}[b]
    \vspace{-3mm}
	\centering
	\includegraphics[width=.9\linewidth]{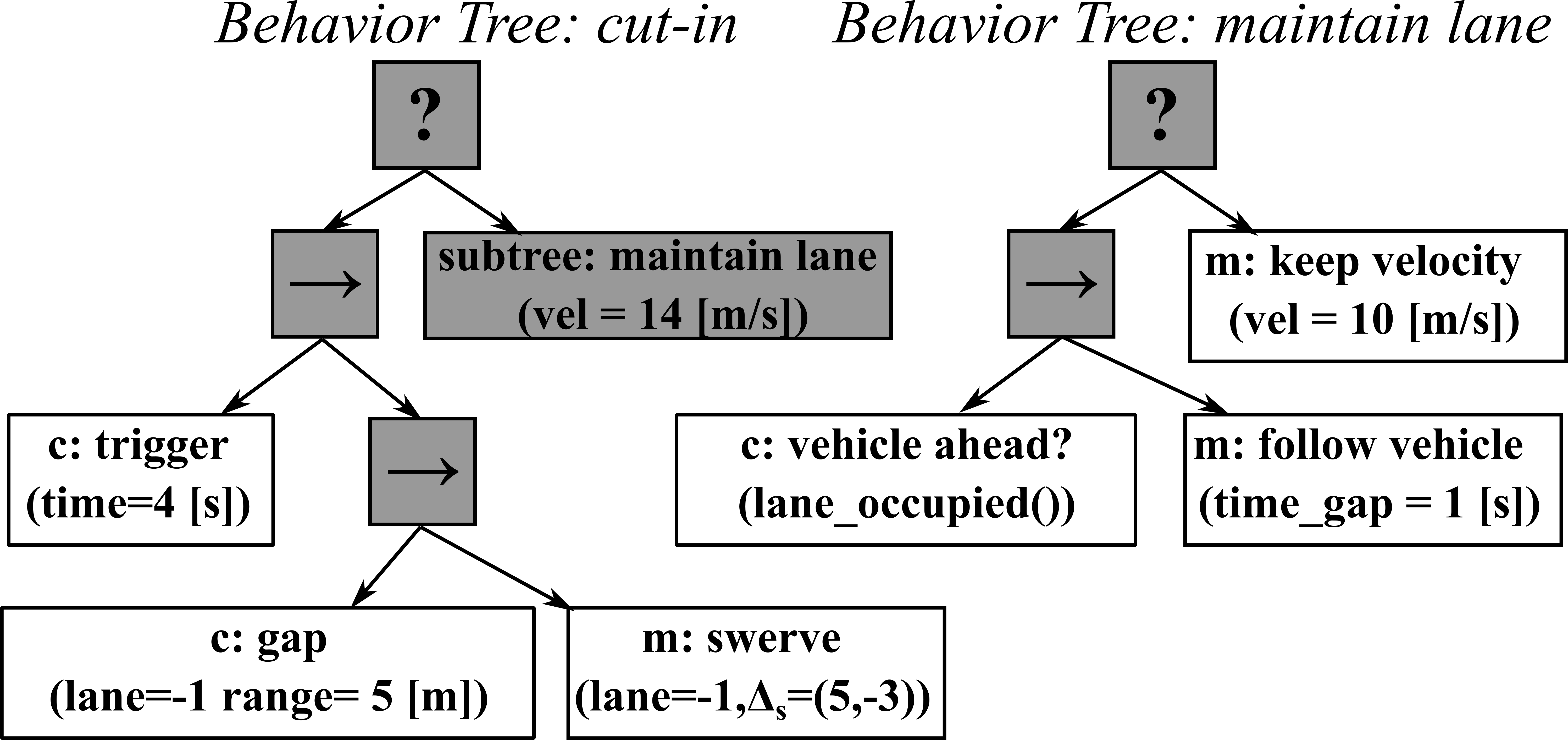}
	\caption{Graphical representation of an example SDV behavior tree structuring the decision-making with conditions (c) and maneuvers (m). '$?$' is the \textit{fallback} operator (short-circuit or), and $\rightarrow$ is the \textit{sequence} operator (short-circuit and).}
	\label{fig:btree}
\end{figure}

\looseness=-1
\parhead{Maneuver Selection}
Maneuver selection is expressed using behavior trees. Their leaf nodes are either (i) conditions to be evaluated (based on the traffic state), (ii) decisions that start (or end) maneuvers, or (iii) references to sub-trees. The inner nodes are control nodes, a.k.a. operators, which are responsible for coordinating the execution of their child nodes. There are three operators. The \emph{fallback} operator commands a sequential execution of its children, left-to-right, and returns success immediately when a child succeeds; otherwise, it executes the next child. It returns failure when none of the children succeeds. The \emph{sequence} operator also commands a sequential execution of its children, left-to-right, but returns failure immediately when a child fails; otherwise, it executes the next child. It returns success when all of the children succeed. The \emph{parallel} operator commands the execution of all children at the same time. The rule for success or failure of the parallel operator is user-defined.

\looseness=-1
\Figref{fig:btree} shows a graphical representation of two example behavior trees, with the left one being the main tree planning cut-in behavior, and the right one being a sub-tree referred to from the main one and performing lane maintenance. The main behavior tree would be assigned to an SDV, e.g., $v_2$ in \figref{fig:cutinscenario}. In each maneuver planning cycle of $v_2$, the main tree is ``ticked'' ($p_i$.4:tick in \figref{fig:flowchart}), i.e., executed, with the local traffic state as context. The execution starts with the root of the main behavior tree and traverses the nodes according to the operator semantics. In our example (\figref{fig:btree}), the execution starts from the fallback operator at the root and proceeds to its child sequence node and then to condition (c:trigger), which tests whether the simulation has been running for 4\,s. If the condition is satisfied, the execution proceeds to the deepest sequence node and then to the condition (c:gap) checking the acceptance distance gap of 5\,m (+-10\%) for a lane change in front of another vehicle to the right (lane\_id=-1). If the gap condition is satisfied, the maneuver node (m:swerve) executes the lane change with a target distance gap of 5\,m and a relative velocity of -3\,m/s ($\Delta_s$=(5,-3)). If any of the two conditions fails, the reference node is executed, triggering the execution of the sub-tree on the right, which implements a simple lane maintenance behavior.

\looseness=-1
A maneuver exposes a set of parameters to control it according to scenario objectives.
We use existing maneuver catalogs\,\cite{sae:2018:j3164,2018:wisedrive:maneuvercatalog} and implement a subset to support the evaluation in \secref{sec:evaluation}: keep velocity, follow vehicle, swerve (used for lane change and swerve-in-lane), merge-in-front, stop, and reverse. Note that these are elemental maneuvers; composite maneuvers are implemented as behavior trees over the elemental maneuvers. For instance, lane maintenance composes velocity keeping, vehicle following, and stopping (for more complex examples see Queiroz\,\cite{queiroz:2022:thesis}).

\looseness=-1
A maneuver node (e.g., m:swerve in \figref{fig:btree}) is represented by a \emph{maneuver configuration} (mconfig in \figref{fig:flowchart}), consisting of the maneuver type (e.g., swerve) and a set of maneuver-specific parameter values, such as the target gap distance and velocity delta for swerve. The behavior tree execution ('tick') is expected to return a maneuver configuration, which is passed to maneuver trajectory planning. 
A given maneuver ends when a condition for a new maneuver is triggered in btree. 

\begin{figure}[h]
    \vspace{-2mm}
	\centering
	 \includegraphics[width=3.5in]{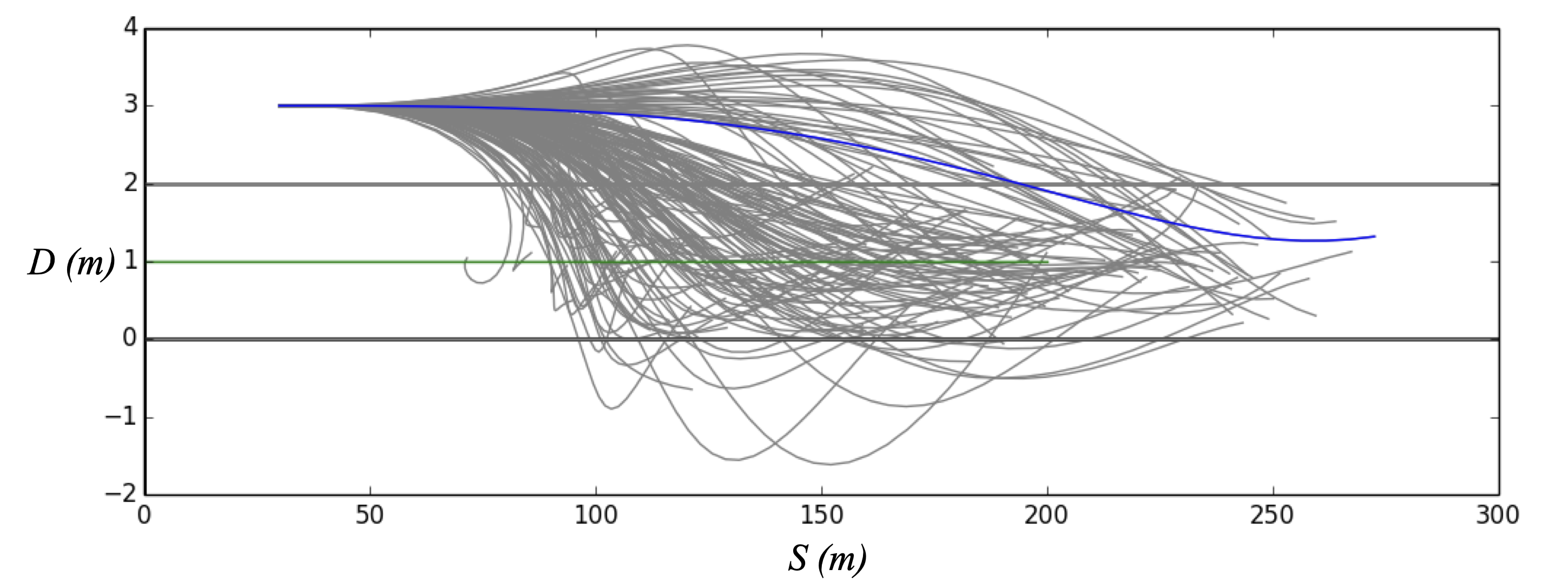}
	\caption{Trajectory planning by the SDV during a cut-in maneuver}
	\vspace{-2mm}
	\label{fig:cutintrajectory}
\end{figure}

\looseness=-1
\parhead{Maneuver Trajectory Planning}
The maneuver trajectory is planned by $p_i$.5:plan\_mtraj in \figref{fig:flowchart} using the maneuver configuration (mconfig) and local traffic (ff\_local\_TS) as inputs. A trajectory is represented by longitudinal $S(t)$ and lateral $D(t)$ position in Frénet frame as functions of time and the trajectory duration $T$ (\ref{eq:traj}). Velocity and acceleration are the first and second derivatives, respectively, yielding the longitudinal and lateral state\,(\ref{eq:frenetstate}):

\vspace{-2mm}
\begin{flalign} 
\textit{Trajectory}&=[S, D, T] \label{eq:traj}\\
\textit{VehicleState}_\text{Frénet}(t)&=[S(t-t_0), \dot{S}(t-t_0), \ddot{S}(t-t_0),\nonumber\\
&D(t-t_0), \dot{D}(t-t_0), \ddot{D}(t-t_0)] \nonumber\\
&\text{  for  } 0\leq t-t_0 \leq T \label{eq:frenetstate}
\end{flalign}

\looseness=-1
The maneuver trajectory planning has three steps: (i) sampling the target states for the maneuver, (ii) generating candidate trajectories, and (iii) selecting an optimal trajectory. Each of these steps is controlled by a set of parameters accessible through the maneuver configuration and allowing testers to realize a particular driving style or misbehavior. The generated trajectories are kept short (2 to 5\,s), but some maneuvers, e.g., vehicle following, are performed over extended periods of time and thus consist of a sequence of trajectories. A behavior tree decides when to start, finish, or abort a maneuver.

\Figref{fig:cutintrajectory} shows an example of trajectory planning for a cut-in maneuver to the right lane. The grey lines are the candidate trajectories eliminated by the optimization step due to feasibility constraints or higher cost. The blue line is the best cut-in trajectory based on motion constraints and the scenario goals specified in mconfig, such as the target gap to ego. The green line is the ego trajectory. The remainder of this section describes each of the three planning steps in more detail.

\subsubsection{Maneuver target sampling (\textnormal{$p_i$.5.1:sample\_targets})}
Each maneuver has its own configurable criteria to define its target state and a time to reach it ($T$). Target sampling requires evaluating the road structure, traffic, and other objects. For example, in the NHTSA pre-crash scenario  `following vehicle making maneuver' scenario, a leading vehicle decelerates to turn right that may end up in a crash with an inattentive following vehicle.

In such scenario, the leading vehicle's target for velocity keeping is to comfortably accelerate to and maintain a specified target velocity, e.g., 16 m/s. The following vehicle's target for vehicle following is to reach and keep a certain target time gap, e.g., 10\,s. 
While defining the maneuver configuration, parameters can be set as a single value or a value range, e.g., a vehicle target speed of exactly 14\,m/s, or within 20\,\% from 14\,m/s. The target sampling step samples multiple values for each range parameter independently and creates a target state set as a Cartesian product over the parameter value sets. The sampling method of choice and the number of samples per parameter are configurable through mconfig. The target state set corresponds to the end points of the trajectories in \figref{fig:cutintrajectory}.

\subsubsection{Trajectory generation (\textnormal{$p_i$.5.2:gen\_traj})} 
Given a target state set, trajectory generation computes a smooth motion profile between the current vehicle state and each target state in the Frénet frame (\figref{fig:cutintrajectory}). We use an approach that plans each trajectory as a pair of quintic polynomials, in longitudinal and lateral direction, respectively, which minimizes jerk to reflect smooth and comfortable driving~\cite{werling:2010:optff}. A quintic polynomial is a jerk-minimal connection between two points $P_0$ and $P_T$, with $p(t)$ as location and $T$ as the motion duration~\cite{1989:takahashi}. More precisely, such a quintic polynomial minimizes the total accumulated jerk over the one-dimensional trajectory:

\begin{equation} \label{eq:jerk}
J_{p,T} := \int_{t=0}^{t=T}  \dddot{p}^{2}(t)dt
\end{equation}

This step generates a trajectory by computing the coefficients of two quintic polynomials, $S(t)$ for the longitudinal dimension as $p(t)$, and $D(t)$ for the lateral direction as $p(t)$, to fit the boundary conditions: the initial state $\textit{VehicleState}_\text{Frénet}(t_0)$ and each of the target states $\textit{VehicleState}_\text{Frénet}(t_0+T)$ from the target-sampling step. This results in a candidate set that respect the maneuver target constraints.

\subsubsection{Optimal trajectory selection (\textnormal{$p_i$.5.3:optimize})}
\begin{figure}[t]
    \vspace{-2mm}
	\centering
		\includegraphics[width=\linewidth]{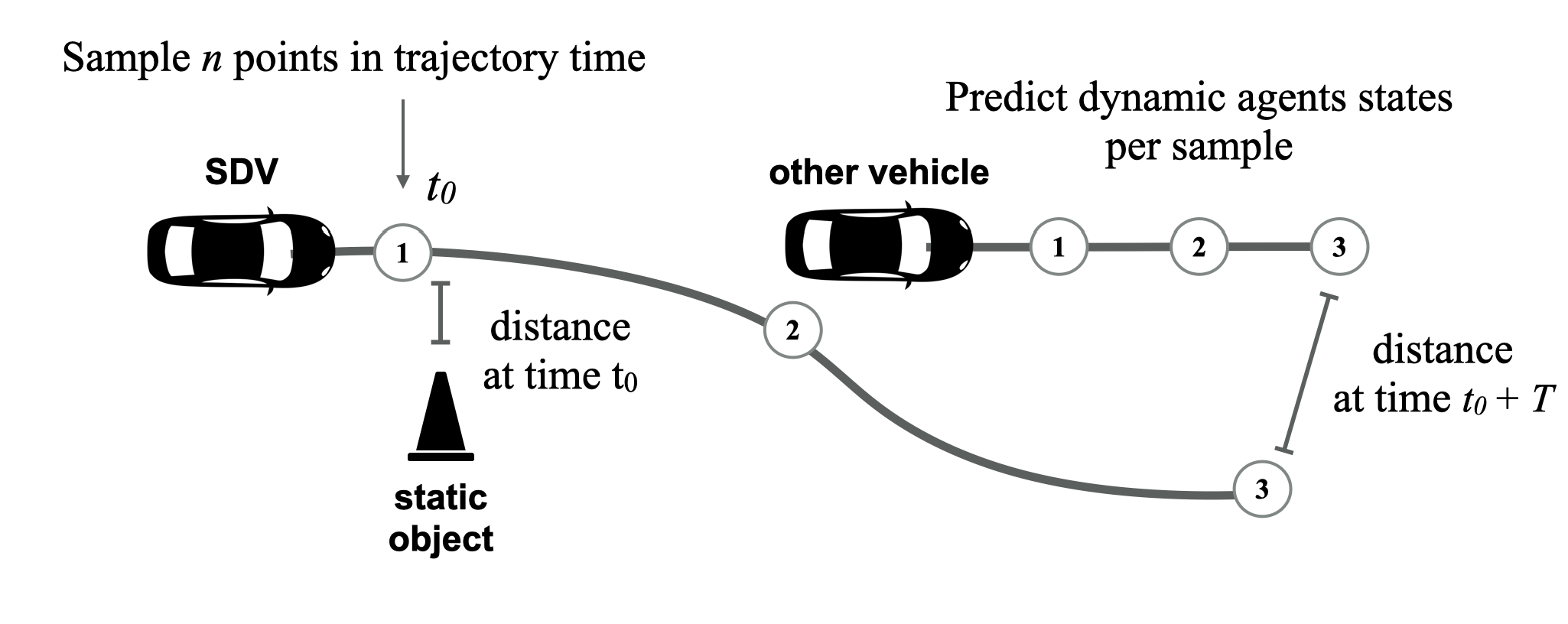}
	 \vspace{-8mm}
	\caption{Checking for collisions with static objects and dynamic obstacles}
	\vspace{-2mm}
	\label{fig:collisioncost}
\end{figure}

This step selects a trajectory that is feasible and optimal with respect to a set of maneuver feasibility constraints and cost functions, which are configured in the maneuver configuration to suit the needs of the test scenario. Feasibility constraints reject trajectories with any collision, direction inversion, lane departure, and exceedance of maximum lateral/longitudinal jerk and acceleration. These are checked by sampling points over the planned and predicted trajectories (e.g., ego), as illustrated \figref{fig:collisioncost}. Note that the optimization step predicts the motion of other vehicles by assuming constant longitudinal velocity in Frénet frame.

Furthermore, the constraints can be configured to fit the scenario objectives. For example, the behavior tree of a $v_2$ (\figref{fig:cutinscenario}) may issue a swerve maneuver with a configuration that disables its collision check to simulate a reckless cut-in.

The remaining candidate set is ranked using a weighted sum of configurable cost functions:
\begin{itemize}
    \item \emph{Time cost} penalizes trajectories longer or shorter than the target time $T$.
    \item \emph{Efficiency cost} penalizes low average velocity.
    \item \emph{Lane-offset cost} penalizes distance from lane center during the entire trajectory.
    \item \emph{Jerk cost} penalizes high longitudinal and lateral jerk over the entire trajectory ($J_{S,T}$ and $J_{D,T}$).
    \item \emph{Acceleration cost} penalizes high longitudinal and lateral acceleration over the entire trajectory.
    \item \emph{Proximity cost} penalizes proximity to obstacles (vehicles, pedestrians, or other objects).
\end{itemize} 
The best trajectory is the lowest-cost feasible one. Weights can be adjusted per behavior tree node according to scenario goals. For example, if a given scenario requires the vehicle to drive too close to ego, the proximity cost weight for ego should be lowered. The resulting trajectory respects realistic vehicle motion, balances conflicting qualities such as progress and comfort, while implementing the scenario goals.

\parhead{Traffic Simulation Execution}
Traffic plans are executed in Traffic Simulation, a process that sets the traffic state of each SDV in TS according to its planned trajectory. It runs at a fixed frequency that is typically an order of magnitude higher than that of an SDVPlanner.
The new trajectories produced by the SDVPlanner processes arrive asynchronously in the traffic plan TP (e.g., $p_i$.6:write). Traffic Simulation retrieves the state of each SDV for the current simulation time from TP, transforms it to the global Cartesian frame of the simulation (TSim.1:ro\_cf), and updates the state of the corresponding SDV in TS (TSim.2:update). Note that updates to TP ($p_i$.6:write) and TS (TSim.2:update) are atomic.

\section{Model Implementation}
\label{sec:implementation}
\noindent\looseness=-1
A reference implementation for the SDV model and tools for running scenarios in simulation are available as part of the open-source project GeoScenario Server. The server parses scenario definitions expressed using Lanelet2 map\,\cite{poggenhans:2018:lanelet2} and the GeoScenario language\,\cite{queiroz:19:geoscenario} extended with the SDV behavior-tree definition format and creates a traffic simulation with the SDV model instances running concurrently. The server is implemented in Python and operates as a co-simulator to be interfaced with the simulation of the ego vehicle, its sensors, and the ADS under test. The implementation also provides a sample integration with an existing simulator, WISE Sim, and an ADS software stack, WISE ADS. The \textit{GSClient} component provides a shared memory interface between the GeoScenario Server and WISE Sim, which runs within the game engine Unreal and provides LiDAR and camera simulation. The high-fidelity dynamics model of the ego vehicle, a Lincoln MKZ, runs as a Robot Operating System (ROS)\,\cite{online:ros} module along with the WISE ADS. The GeoScenario Server can be integrated into any other simulation environment, simply by customizing GSClient for the new environment.

\section{Evaluation}
\label{sec:evaluation}
\noindent 
We evaluate the SDV model in terms of design effectiveness, realistic vehicle motion, practical applicability for scenario-based ADS testing, and scalability. The following research questions guide our evaluation:

\begin{itemize}
\item \textbf{RQ1}: Can realistic and interactive scenarios for ADS testing be effectively modeled and executed via SDV models?
\item \textbf{RQ2}: Can SDV models generate realistic vehicle motion?
\item \textbf{RQ3}: Can the use of SDV models improve the effectiveness of scenario-based testing of a real ADS?
\item \textbf{RQ4}: How does the model performance scale with traffic density?
\end{itemize}

\subsection{Effective Scenario Development (RQ1)}
\label{sec:rq1}
\noindent
We evaluate the effectiveness of scenario development using the SDV model by analyzing how the model improves GeoScenario as the baseline DSL to design and execute test scenarios from a catalog using three metrics:

\begin{itemize}
\item (i) \textit{Expressiveness}: Given a set of scenarios, we classify them as follows: we assign \textit{success} (S) when all behaviors are successfully expressed with no limitations, \textit{partial} (P) when the behaviors for at least one variation of the scenario can be expressed, or \textit{failure} (F) otherwise.

\item (ii) \textit{Execution accuracy}: After running a simulation, we classify the degree to which scenarios are correctly executed according to NHTSA description: \textit{success} (S) when all vehicles behave as expected and the scenario objective is achieved; \textit{partial} (P) when at least one variation of the scenario succeeds; and \textit{failure} (F) otherwise.

\item (iii) \textit{Reuse}: We quantify reuse in a scenario based on the \emph{internal reuse level}~\cite{frakes:1996:reusemetrics}. Given a scenario containing a set of behavior trees (higher-level items), the metric is defined as $M/L$, where $M$ is the number of nodes (lower-level items) that are used more than once (i.e., used also in behavior trees of other scenarios) and $L$ is the total number of nodes in the set of behavior trees. This metric assumes values between 0 and 1 and represents the percentage of internal reuse. We also compute the internal reuse level accounting for only the nodes that are actually executed in a successful simulation.

\end{itemize}

\looseness=-1
Since the SDV model extends the capabilities of GeoScenario, we use the latter as the baseline\,\cite{queiroz:19:geoscenario}.
We focus on safety-critical scenarios and, specifically, we use the Pre-Crash Scenario Typology from NHTSA\,\cite{nhtsa:2007:crashtypology}. 
These interactive and realistic scenarios can challenge the ADS capabilities in crash avoidance and they are commonly used as a reference for ADS validation in other projects\,\cite{2020:waymosafety,dosovitskiy:2017:carla}. We filter the original set for scenarios with vehicle-to-vehicle interactions, resulting in 18 scenarios (Table\,\ref{tab:scenarios}).

\looseness=-1
We design each scenario using a combination of the original GeoScenario and multiple instances of SDV models with their respective behavior trees and maneuver configurations. The original NHTSA set is based on reported events between human-operated vehicles, but we assume that one of the vehicles is ego, operated by the ADS (similar to how Waymo adapts NHTSA scenarios as tests~\cite{2020:waymosafety}). Ego's goal is to drive through the scenario (from start to goal point) and avoid a collision. The goal of an SDV is to interact with ego using target parameters defined by the tester,  e.g., achieving a certain time gap before braking. The overall scenario goal is to replicate the pre-crash events as described by NHTSA, leading to a crash or a near-crash. If execution differs by either a safe outcome (vehicles never interact or interact differently than intended) or another type of crash, the scenario execution fails. After modeling the scenarios, we simulate them in the reference implementation (\secref{sec:implementation}).

\looseness=-1
As part of the comparison of expressiveness with the baseline, we classify the type of SDV behavior required in each scenario as \textit{static} or \textit{dynamic} with respect to three elements: \textit{path shapes}, \textit{speed profiles}, and \textit{behavior triggers}. Behavior triggers are conditions triggering the required changes in paths and speed profiles during the scenario (\tabref{tab:scenarios}). Scenarios that involve static behavior for all three elements, i.e., fixed paths and speed profiles for each SDV and their starting triggers, can be easily designed with predefined trajectories from start to finish and do not benefit significantly from a dynamic model (stat,stat,stat in \tabref{tab:scenarios}). Scenarios that require dynamic behaviors, but the behaviors can be expressed as sets of static paths and velocity profiles with dynamic triggers to select among them (stat,stat,dyn in Table \ref{tab:scenarios}), can still be modeled using predefined trajectories with reasonable effort. Finally, scenarios that require dynamic path or velocity profile or both (dyn,stat,*; stat,dyn,*; and dyn,dyn,* in Table \ref{tab:scenarios}) are impractical to be modeled using predefined trajectories, but are enabled by the proposed SDV model. For example, the cut-in scenario has a continuous space of paths and speed profiles, and a dynamic trajectory needs to be planned based on the ego behavior, which may vary from execution to execution. We note that using the NHTSA descriptions of the scenarios as a source, many scenario variants are possible. Our classification is based on the minimal behavior required to reproduce the critical event occurring immediately prior to a crash as described by NHTSA; however, added elements, such as additional vehicles, might change the static classification to a dynamic one, but not the other way.

\textit{\textbf{Results:}}
Due to limited space, we focus on the main findings here. The full list of scenarios is in the online repository.\footnote{\href{https://github.com/rodrigoqueiroz/geoscenarioserver}{https://github.com/rodrigoqueiroz/geoscenarioserver}}

\looseness=-1
\textit{Expressiveness}:
All 18 scenarios except for one variant of \#17 are successfully expressed using the SDV model.
We identify 14 scenarios (78\%) that depend on dynamic path or velocity profile, or both, and thus are impractical for the baseline. For instance, a vehicle leaving a parking position in scenario \#17 must start this maneuver only when ego is approaching and adjust its trajectory, in one of the variants, to merge ahead of ego. While the vehicle must challenge the ADS, an unavoidable lateral crash into ego would not be useful as a test scenario. To achieve the scenario goal, the vehicle must be able to generate a trajectory relative to ego's motion at run time. The same requirement applies to all lane-change scenarios (\#16-\#19). For crossing-path scenarios \#30 and \#31, the velocity profile must be dynamically planned. The SDV models enable us to successfully express these dynamic behaviors, which are infeasible with the baseline, resulting in a higher expressiveness. One variant of Scenario \#17 ``Parked Vehicle SD'' requires the parked vehicle to join traffic by making a U-turn, and this maneuver is currently not supported by the implementation of trajectory generation.
	
\looseness=-1
A total of four scenarios (22\%) require only static trajectories (stat,stat,* in Table \ref{tab:scenarios}) and thus can be designed with the baseline. For instance, in the rear-end scenario \#25 both path shape and speed profile can be generated offline and expressed as predefined trajectories with only a trigger to activate the deceleration as ego approaches. In such examples, the SDV model does not increase expressiveness. However, it adds two advantages: (i) conciseness, by defining the scenario at a higher level of abstraction using target parameters instead of detailed trajectories, and (ii) flexibility, by allowing the scenario to be replicated in different road geometries without changing the behavior definition. 

\textit{Execution}: In 17 scenarios, vehicles perform as expected, and the scenario ends with a crash or near-crash as described in the NHTSA report. The performance deviates from the design in the scenario \#16 ``Vehicle(s) Turning – Same Direction.'' The assigned behavior requires that vehicles perform a maneuver that violates the legal road-network connectivity. 
Since the current implementation relies on the Lanelet map to constrain the driving space, the map requires an adaptation to execute the scenario correctly.

\looseness=-1
\textit{Reuse}: The composable nature of behavior trees allows us to reuse most of them, i.e., use each tree in two or more scenarios, since there is significant commonality in the driving task for the different scenarios. In most scenarios, vehicles start by performing normal lane maintenance until an unexpected event occurs, such as a risky behavior of another vehicle. The differences among scenarios emerge in such events and are usually modeled at the highest levels of the main behavior tree for the given scenario. We call them the ``\textit{scenario-trees}.'' The remaining tasks are reusable and performed using ``\textit{sub-trees}" (e.g., performing a lane-change). This reuse pattern is not part of the original behavior-tree concept, but it has emerged during this experiment when trying to maximize reuse. In some instances, a simple overriding of parameters for conditions or maneuvers during the sub-tree composition is sufficient to adapt the behavior from one scenario to another and achieve the scenario objective with 100\% reuse (see \textit{Internal Reuse Level} in Table~\ref{tab:scenarios}). Overall, the average internal reuse level (weighted by the size of behavior trees in each scenario) is 0.93 for all nodes, and 0.81 for executed nodes.

\looseness=-1
The experience modeling and running NHTSA scenarios reveals how effective the SDV model can be in ADS scenario development. The model enables expressing highly dynamic behaviors, fosters reuse, and can successfully execute most scenarios in simulation. Vehicle interactions involving lane changing, merging, and crossing paths are severely limited or impractical using the baseline of predefined trajectories. Thus, such interactive scenarios benefit most from the SDV model. The limitations we identify are due to missing underlying maneuvers (such as a U-turn) or the map constraints that prevent certain vehicle movements. We will address them in future work.


\begin{table}[t]
\centering
\caption{Scenarios and performance}
\label{tab:scenarios}
\tabcolsep=0.05cm
\begin{tabular}{lcp{3.7cm} | lll | llll}
\toprule
{ID} &Group &Scenario &\rotatebox{90}{Path Shape} & \rotatebox{90}{Speed Profile} & \rotatebox{90}{Behavior Trigger} &\rotatebox{90}{Expressiveness} &\rotatebox{90}{Execution} &\rotatebox{90}{IRL}&\rotatebox{90}{IRL exec}    \\
\midrule
4 &CP &Running Red Light                &stat	    &dyn	    &stat   &S &S   &0.83 &0.60 \\
5 &CP &Running Stop Sign                &stat	    &dyn	    &dyn    &S &S   &1.00 &1.00 \\
15 &B &Backing Up \par Into Another Vehicle&stat	&stat	    &dyn    &S &S   &0.91 &0.60 \\
16 &LC &Turning SD                      &dyn	    &dyn	    &dyn    &S &S*  &0.87 &0.63 \\
17 &LC &Parking SD                      &dyn	    &dyn	    &dyn    &P &P   &0.84 &0.71 \\
18 &LC &Changing Lanes SD               &dyn	    &dyn	    &dyn    &S &S   &0.89 &0.79 \\
19 &LC &Drifting SD                     &dyn	    &dyn	    &dyn    &S &S   &0.79 &0.71 \\
20 &OD &Making Maneuver OD              &dyn        &dyn        &dyn    &S &S   &0.90 &0.84 \\
21 &OD &Not Making Maneuver OD          &dyn	    &dyn	    &dyn    &S &S   &0.76 &0.50 \\
22 &RE &Following Vehicle \par Making Maneuver&dyn  &dyn        &dyn    &S &S   &1.00 &1.00 \\
23 &RE &Lead Vehicle Accelerating       &stat	    &stat	    &dyn    &S &S   &0.90 &0.75 \\
24 &RE &Lead Vehicle at Lower Speed     &stat	    &stat	    &stat   &S &S   &1.00 &1.00 \\
25 &RE &Lead Vehicle Decelerating           &stat	&stat	    &dyn    &S &S   &0.90 &0.75 \\
27 &CP &Left-Turn Across Path/OD\par at SJ  &stat   &dyn	    &dyn    &S &S   &0.90 &0.75 \\
28 &CP &Vehicle Turning Right at SJ         &stat   &dyn    	&dyn    &S &S   &0.99 &0.94 \\
29 &CP &Left-Turn Across Path/OD\par at NSJ &stat   &dyn	    &dyn    &S &S   &0.98 &0.93 \\
30 &CP &Straight Crossing Paths at NSJ      &stat   &dyn	    &dyn    &S &S   &0.94 &0.81 \\
31 &CP &Vehicle Turning at NSJ              &stat   &dyn	    &dyn    &S &S   &0.94 &0.81 \\
\bottomrule
\end{tabular}
\bigskip

Acronyms: B: Backing up, CP = Crossing Paths, LC = Lane Change, OD = Opposite Direction, RE = Rear-end, SD = Same Direction, SJ = Signalized Junction, NSJ = Non-Signalized Junction. 
Path Shape, Speed Profile, and Behavior Trigger are requirements for vehicle behavior that can be static (stat) or dynamic (dyn).
Expressiveness and Execution show the degree in which a scenario is modeled and correctly executed, respectively (S=successfully, P=partially, F=Failed). The Internal Reuse Level (IRL) is computed with all Behavior tree nodes, and only for nodes that are executed in the simulation (IRL exec).
*Scenario \#16 required a map adaptation to perform correctly.
\vspace{-2mm}
\end{table}

\subsection{Vehicle Motion (RQ2)}
\label{sec:rq2}
\begin{figure}[t]
	\centering
	 \includegraphics[scale=0.15]{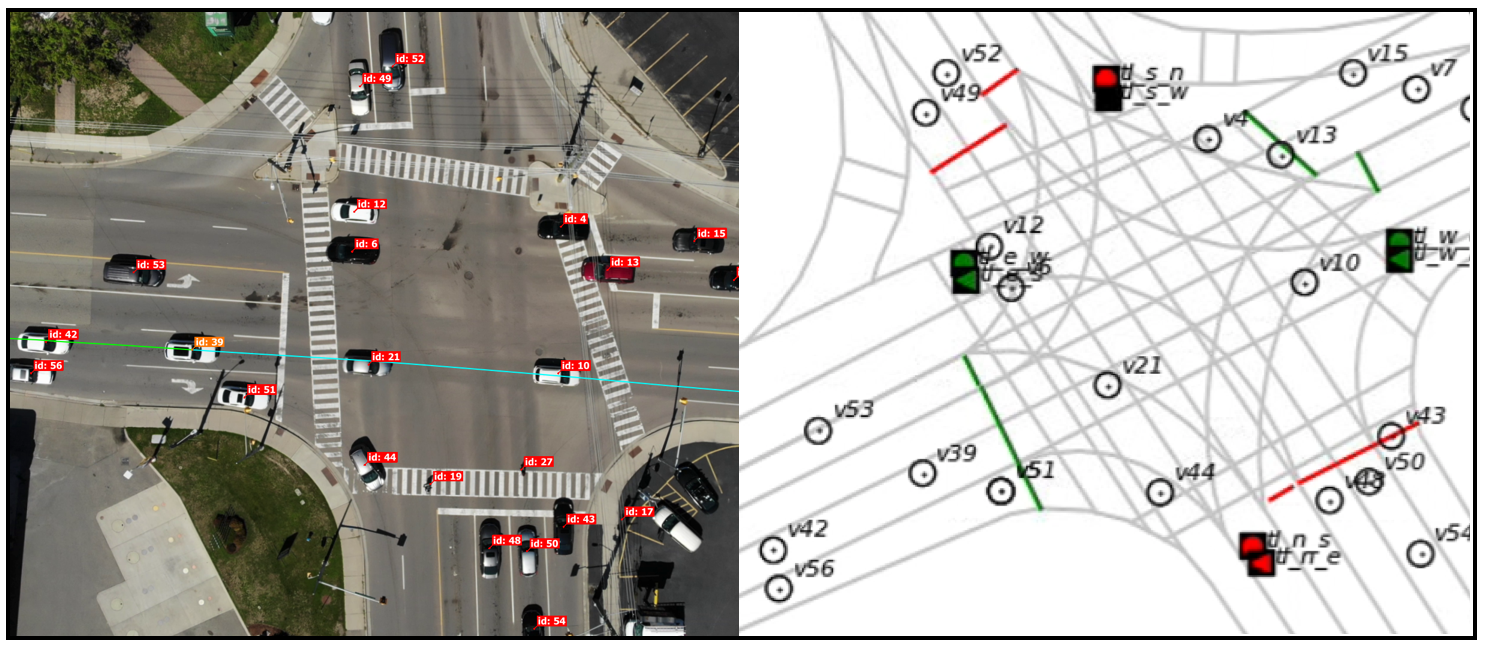}
	\caption{A snapshot of the signalized intersection used for experiments and its corresponding simulation on the right.}
	\vspace{-3mm}
	\label{fig:dataset}
\end{figure}

\noindent\looseness=-1
As the primary goal is to simulate human-operated vehicles, a good model must reflect  the  human-driving  behavior and how vehicles move in naturalistic traffic conditions. To evaluate the motion realism, we use SDV models to replicate scenarios collected from urban traffic  and compare their behavior with real vehicles. It is unreasonable to expect SDV models to drive exactly like the empirical vehicle, since not even humans drive equally. However, our model is designed to be highly configurable and adapt to different driving styles. With the proper configuration in the calibration process, we expect that SDV models can approximate the behavior of the empirical vehicles to a high degree given the same environment conditions. We use data from a busy signalized intersection during mid-day traffic in Waterloo, Canada, which is part of the Waterloo Multi-Agent Traffic Dataset \cite{online:wisetrafficdataset}. The ``birds-eye" video was collected using a drone and processed to label and track pedestrians and vehicles (Fig.~\ref{fig:dataset}). 

\noindent
This experiment follows four steps:
\begin{enumerate}
\item \textit{Data preparation:} We classify the vehicle trajectories in the dataset into five scenario types based on the main maneuver they represent: (i) vehicle crossing intersection unconstrained (free), (ii) vehicle stopping (red light), (iii) vehicle resuming driving (green light), (iv) vehicle following a lead through the intersection (follow), and (v) vehicle partly following a lead when the lead merges or leaves mid-scenario (free/follow). In cases where a vehicle stops at a signal light, we split the trajectory into two scenarios, namely (ii) and (iii), in order to eliminate the waiting state. Each such classified vehicle trajectory  represents an individual experimental trial.

\item \textit{Test generation}: For each classified vehicle trajectory, we identify the traffic conditions that may affect how the vehicle is driving, e.g., signal light states and all other vehicles and pedestrians that may affect it, to be reproduced in simulation. Each classified vehicle trajectory is used as a reference vehicle for a single test. We generate a new GeoScenario test replacing the reference vehicle with an SDV model instance with a standard-driver behavior tree and using the same start state (velocity and position in the intersection), and replicate the traffic conditions to ensure the driving task is influenced by the same factors. The standard-driver behavior tree is capable of performing each of the five maneuvers. We also assign a route goal to the model based on the last known position of the empirical reference vehicle to ensure the simulated vehicle will navigate the intersection towards the same exit lane. All other relevant empirical vehicles and pedestrians are included in the test as agents with predefined trajectories, and the signal light phases are also replicated. We generate 100 test scenarios and manually review the correctness of the identified traffic conditions.

\item \textit{Calibration}: 
While each simulated reference uses the same standard-driver behavior tree, it needs a behavior-tree configuration to replicate the driving style of its empirical counterpart.  
We use a set of rules to automatically analyze each empirical reference trajectory and generate a configuration for it by extracting a set of high-level driving-style parameter values and value ranges, including maximum and average velocities, lateral displacement on the lane, stopping distance to target, reaction times, and time gap to other vehicles. We adjust the SDV parameter ranges to target similar values.

\item \textit{Simulation}: We run two simulations per scenario using the SDV model, one with a default configuration before the calibration and another one after the calibration, and export the resulting trajectories as a discrete set of the vehicle states in the simulation frequency at 30\,Hz. The default configuration uses nominal naturalistic driving parameters, such as zero offset from the lane centerline and a time gap range of 1.8..2.2\,s~\cite{2018:wisedrive:motioncontrol}. 
\end{enumerate}

The SDV performance is assessed using a measure of distance between the simulated trajectory $T_1$ and the empirical reference trajectory $T_2$, which takes into account both their spatial and temporal characteristics. The shorter the distance, the more similar the motion behavior of the simulated and the empirical vehicle. We use the spatio-temporal Euclidean distance (STED)~\cite{nanni:2006:sted}, which represents the average Euclidean distance between positions of the respective vehicles, $T_{1}(t)$ and $T_{2}(t)$, along their respective trajectories $T_{1}$ and $T_{2}$, over the interval $l$ in which both trajectories exist:

\vspace{-2mm}
\begin{equation}\label{eq:sted}
d_{\textit{STED}}(T_1,T_2) = 
\frac{\int_{l}^{}  d(T_{1}(t) , T_{2}(t))\,dt}{|l|}
\end{equation}

\begin{figure}[!ht]
	\centering
	 \vspace{-2mm}
	 \includegraphics[scale=0.50]{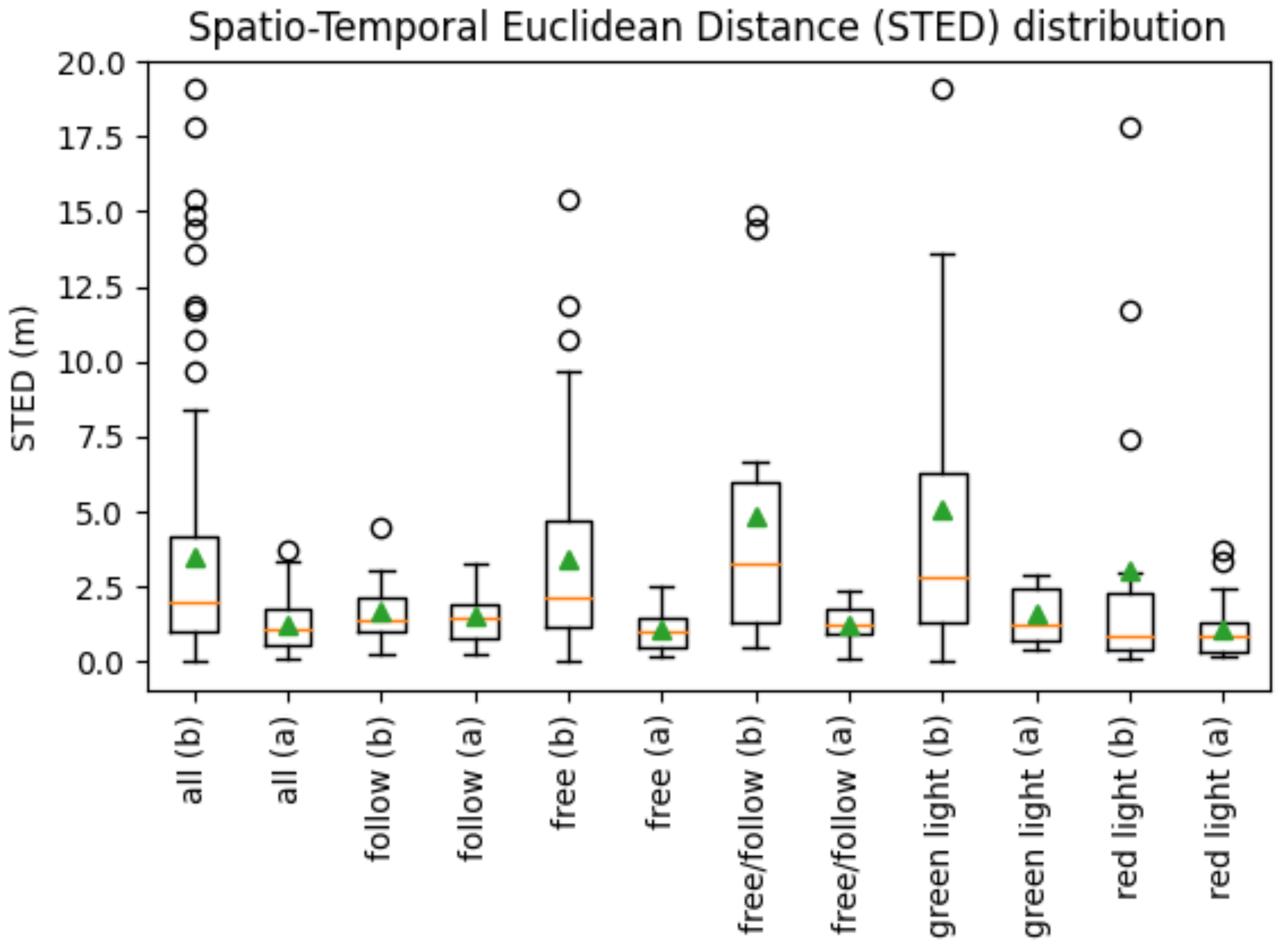}
	\vspace{-2mm}
	\caption{Performance (Eq.~\ref{eq:sted}) for all scenarios and per type, before (b) and after (a) calibration, measured using STED in meters. Orange lines represent medians, and green triangles represent averages.}
	\label{fig:boxplot}
\end{figure}

\textit{\textbf{Results:}}
\Figref{fig:boxplot} shows the distribution of STED before and after calibration per scenario type.
The majority of simulated trajectories are already fairly similar to their empirical reference even before the calibration with an average STED of 4.27\,m.
A review of the simulated trajectories shows a similar decision making patterns, such as reacting to traffic lights and vehicles ahead, to the empirical ones. However, the main differences are observed in the speed profiles, lateral placement on the lane, time gaps, and various delays and reaction times, all indicative of different driving styles. The calibration brings the simulated trajectories significantly closer to their empirical counterparts: average STED for all 100 scenarios reduces from 4.27\,m to 1.24\,m. At an individual level, calibration improves the performance in 82 scenarios. Although the performance is worse for 18 scenarios, it is only slightly worse for 16 of them, with less than 1\,m deterioration. Only two scenarios deteriorated more significantly, by 1.4\,m and 1.9\,m. The latter deviation is due to an erratic driving style of the empirical reference vehicle, which accelerates hard when resuming driving on green and then decelerates for no apparent reason. Such erratic behavior could be replicated by a dedicated maneuver.

\begin{figure}[!t]
    \vspace{-2mm}
	\centering
    \includegraphics[width=1.0\linewidth]{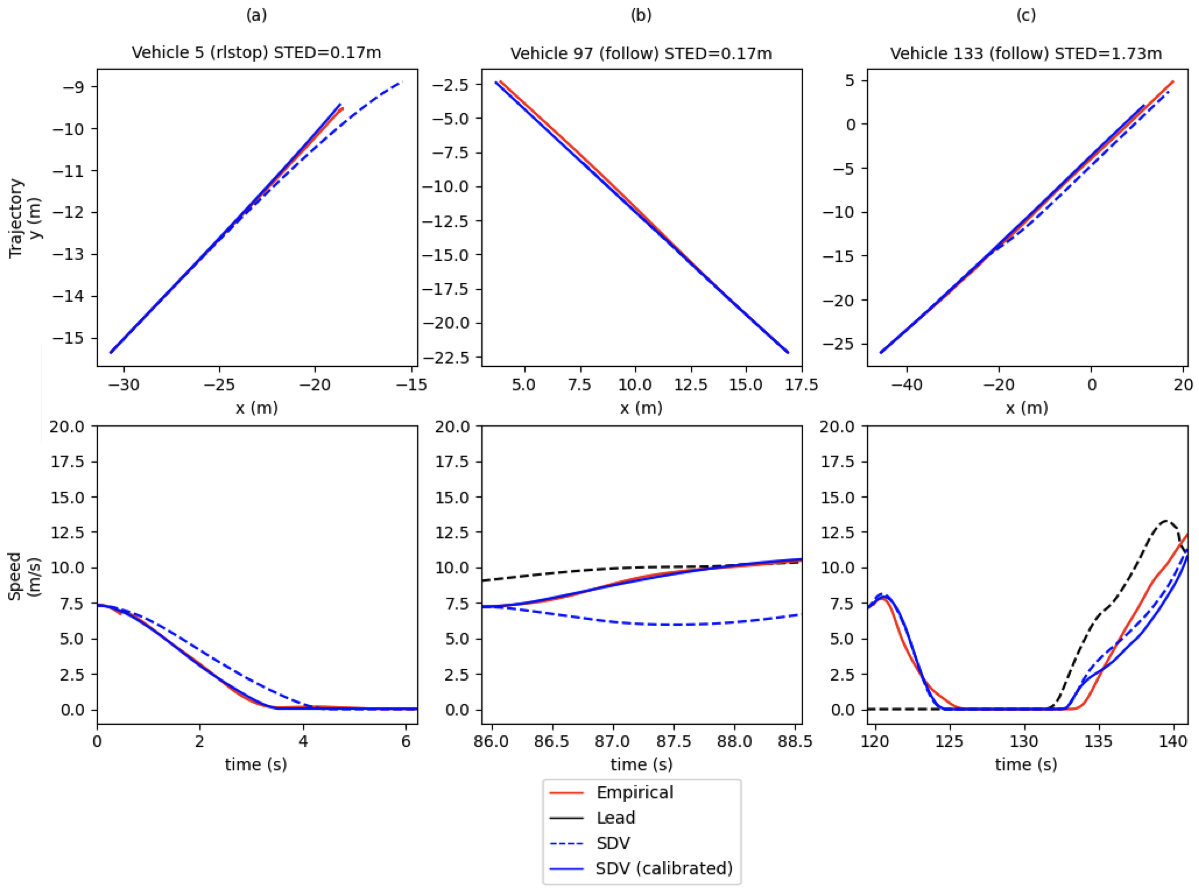}
    \vspace{-6mm}
    \caption{Paths and speed profiles for three sample scenarios. Empirical vehicles in red; SDV models in dashed blue (before calibration) and solid blue (after calibration). Eq.~\ref{eq:sted} defines performance using the spatio-temporal Euclidean distance (STED).}
    \label{fig:results_trajectories}
\end{figure}

\Figref{fig:results_trajectories} shows the paths and speed profiles of sample individual scenarios.
Plot (a) shows the reference vehicle 5 reacting to a red light. The path before calibration shows the simulated vehicle stop at the stop line, but the empirical vehicle stops about 2.5\,m before the line. After calibration, both the simulated and empirical paths match up almost perfectly, with an STED of 17\,cm, and a maximum distance of 31\,cm.
The calibrated speed profile also closely matches the empirical one.
Plot (b) shows vehicle 97 crossing the intersection southwards, while already following a lead vehicle. The black dashed line shows the lead vehicle’s speed profile, which is fairly constant throughout the scenario. The initially slower reference vehicle accelerates to match the lead’s speed. The calibration improves the default configuration to match the more aggressive time-gap of the empirical vehicle, resulting in closely matched speed profile and reducing the STED from 2.37\,m to 17\,cm. In rare cases, the calibration does not improve performance, as shown in plot (c). A vehicle approaches the intersection with a red light and an already stopped vehicle ahead. The reference vehicle can resume driving on the green light, but needs to keep distance from the lead vehicle. The simulated vehicle resumes with a smaller delay compared to the empirical one.

\looseness=-1
In summary, SDV models can closely reproduce the behavior of human-operated vehicles under the same traffic conditions. The model calibration can address varying driving styles and significantly increase the similarities in the trajectories. In some scenarios, such as in \figref{fig:results_trajectories} (a), the simulated trajectory after calibration is in essence indistinguishable from the empirical one, with maximum difference of 31\,cm. In some scenarios the human behaves unexpectedly, however, and the current automatic calibration process cannot replicate such behaviors, but they could be modeled in the behavior trees as additional maneuvers.

\subsection{Application (RQ3)}
\label{sec:rq3}
\noindent
We run an in-depth case study to evaluate how the model performs in a real ADS testing environment and answer RQ3. We choose the cut-in lane change NHTSA scenario (\#18 in Table \ref{tab:scenarios}) to test an actual ADS software as the subject system. In this scenario, a vehicle changes lanes at a non-junction and merges closely in front of the ego traveling in a adjacent lane in the same direction. Cut-in maneuvers from other drivers pose challenges to the ADS and if not handled properly can lead to crashes. Thus, they  represent an important test case.

\looseness=-1
The test aims to evaluate the impact of key vehicle interaction parameters, such as relative velocity and gap, on the likelihood and crash severity. The non-deterministic behavior of the subject ADS makes simulating this type of scenario challenging, however. Reaching the desired test parameter values while performing realistic vehicle interactions requires a reactive model, capable of adapting and re-planning trajectories as the scenario unfolds.

The case study has an explorative nature, with the objective to generate practical insights of applying the SDV model to test a real ADS, including identifying potential limitations. 

\subsubsection{System under test}
\label{sec:subjectsystem}
We test \textit{WISE ADS}, developed at the University of Waterloo.\footnote{\url{https://uwaterloo.ca/waterloo-intelligent-systems-engineering-lab/projects/wise-automated-driving-system}} The ADS software consists of a set of ROS modules implementing object-detection and tracking, occupancy and high-definition mapping, localization and state estimation, maneuver and trajectory planning, and control. The software can operate a Lincoln MKZ Hybrid, equipped with a drive-by-wire interface and a suite of LiDAR, camera, GPS, and inertial sensors, in automated mode at SAE level 3 on public roads in Waterloo. We test the ADS software in simulation, using WISE Sim with the GeoScenario Server implementing the SDV model (see \secref{sec:implementation}).

\subsubsection{Test scenario}
\label{sec:execution}

\looseness=-1
The cut-in behavior is expressed as a behavior tree similar to \figref{fig:btree} and assigned to an SDV model instance. According to this behavior tree, the vehicle must reach a certain acceptance (rear) gap before performing the cut-in maneuver and then achieve a certain target (rear) gap to ego. The behavior tree calls the standard-driver behavior tree to maintain its current lane, parameterized with a target speed of 14\,m/s (+-10\%), which is slightly higher than the road speed limit. The simulation plans candidate trajectories by sampling 6 target velocities from this target range (uniformly, by default). After a delay of 4\,s to allow the vehicle to pick up pace, it starts checking for the acceptance distance gap of 5\,m (+-10\%) for a lane change to the right (lane=-1), on which ego drives at the road speed limit. Once the acceptance gap is satisfied, the lane change is triggered, with a target distance gap of 5\,m and a relative velocity of -3\,m/s ($\Delta_s$=(5,-3)). The experiment repeats the scenario with different combinations of parameters to evaluate how ego handles a variety of cut-in trajectories and find configurations that are more likely cause a crash.

\begin{figure}[t]
\centering
    \includegraphics[width=1.\linewidth]{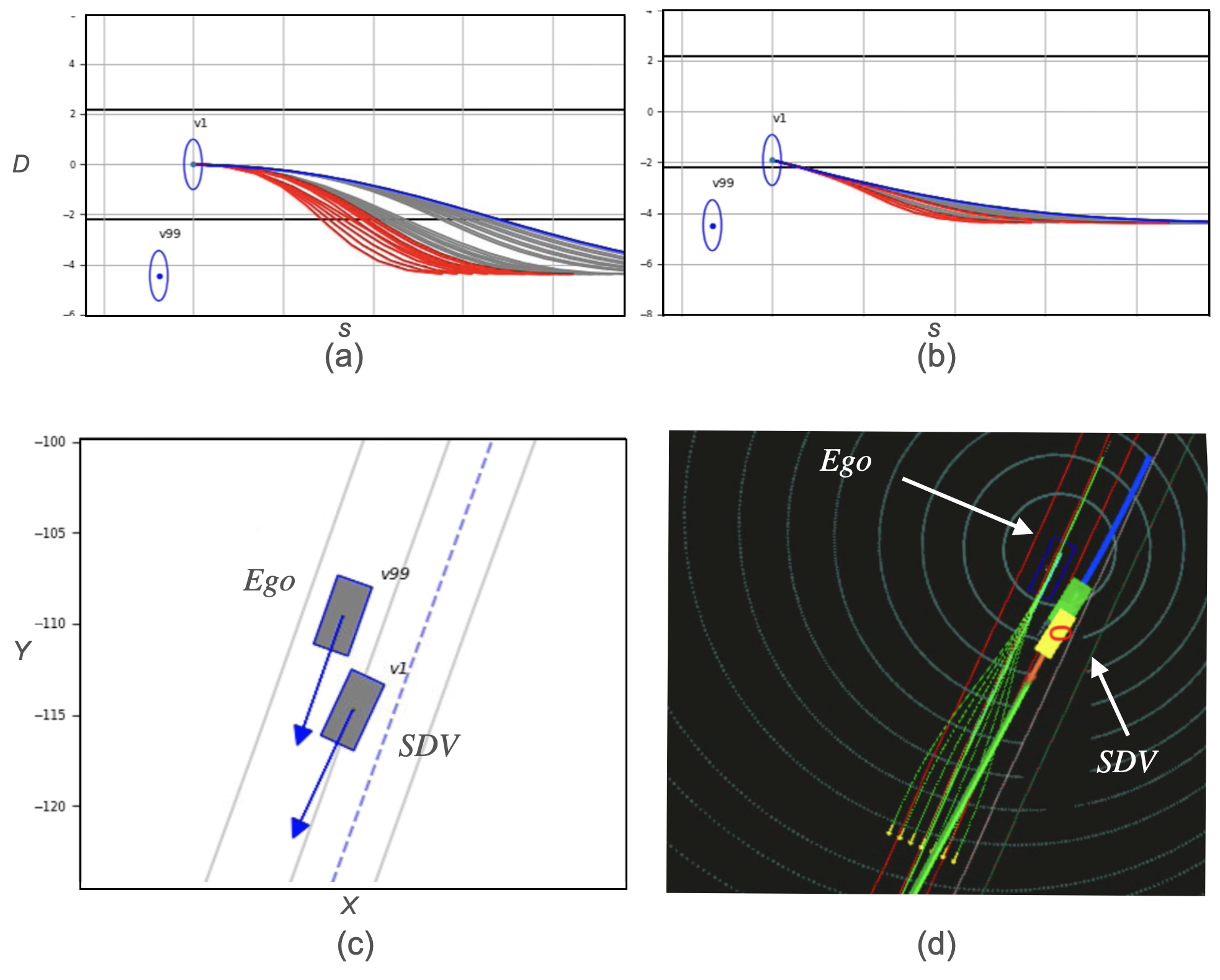}
    \vspace{-4mm}
    \caption{One of the simulation scenarios that results in a crash; in (a) and (b), the SDV trajectory generation in Frénet Frame targeting ego at two different moments (optimal trajectory in blue and infeasible ones in red); in (c) the SDV simulation view in Cartesian coordinates; and in (d) the ADS perception (circles represent the lidar simulation, with the ego located at their center)}
    \vspace{-3mm}
    \label{fig:sim_rviz_server}
\end{figure}

\textit{\textbf{Results:}}\label{sec:results}
As expected, more aggressive cut-ins (shorter acceptance distance gap, shorter target distance gap, and lower target velocity) are more likely to cause collisions, but the response of the ADS to different parameter combinations of the cut-in maneuver is non-obvious (see Table~\ref{tab:simresults}). Scenarios \#7 and \#8 are parameterized with the same short acceptance gap $\Delta d_a$=2\,m and the same target relative velocity $\Delta v_t$=-5\,m/s, but  \#8 has a smaller target distance gap, $\Delta d_t$=-5\,m, compared to $\Delta d_t$=-2\,m for \#7. As a result, \#8  ends in a collision. Note that $\Delta d_t$ and $\Delta v_t$ are planned relative to the predicted ego location at the end of the cut-in maneuver, assuming ego continues at a constant velocity. Thus, although a negative $\Delta d_t$ would guarantee a collision if ego maintained its velocity, ego is likely to brake and thus a negative $\Delta d_t$ does not necessarily result in a collision. Scenarios \#9-11 use a larger acceptance gap, with $\Delta d_a$=5\,m. As a result, although \#9 has the same target parameters as \#8, a collision is avoided, since the larger acceptance gap gives ego more time to react. Increasing the target deltas in \#11 results in a collision, however. Figure~\ref{fig:sim_rviz_server} shows scenario \#8 with the SDV's trajectory generation (a-b), its ground-truth perspective (c), and the ADS’s internal perception of the scenario (d). The ADS detects the SDV (yellow bounding box), and the ADS’s tracker predicts the SDV’s future trajectory (bold green line) as in conflict with the ego’s lane. Although ego initiates an emergency stop, the rear-end collision is not avoided. 

\begin{table}[b]
\centering
\caption{Simulation parameters for SDV behavior and results}
\label{tab:simresults}
\tabcolsep=0.05cm
\begin{tabular}{l|rrr|rcrrl}
\toprule
 & \multicolumn{3}{c|}{SDV Config} & \multicolumn{5}{c}{Observed} \\
\midrule
\# &$\Delta d_a$ &$\Delta d_t$ &$\Delta v_t$ &$\Delta d_a$ &Coll. & $v_{SDV}$ & $v_{Ego}$ &maneuver\\ 
\midrule
7  &2 &-2  &-5  &2.07   &n &- &-            &-\\
8  &2 &-5  &-5  &2.05   &y &10.89 &13.16    &emergency stop\\
9  &5 &-5  &-5  &5.49   &n &- &-            &-\\
10 &5 &-5  &-10 &5.50    &n &- &-            &stop\\
11 &5 &-10 &-10 &5.60   &y &7.60 &12.15   &-\\
\bottomrule
\end{tabular}
\vspace{-1mm}
\end{table}

This experiment demonstrates how the SDV model can be used with a real ADS to search for scenarios and parameters where the system may not be able avoid a collision. We found that using another SDV instance as placeholder for ego enables a rapid iterative development of test scenarios. The iterations are needed to ensure the correct behavior of the cutting-in vehicle and select reasonable ranges of test parameters, before running the more time-consuming simulation with ego controlled by the ADS. Finally, the experiment results also highlight the importance of being able to plan the SDV maneuver trajectories dynamically and influence their shape via parameters.

\subsection{Scalability (RQ4)}
\label{sec:rq4}
\noindent
We evaluate the SDV model scalability to see if it can support scenarios with heavy traffic. To support such scenarios, the model must be able to scale traffic density without any significant degradation of the simulation performance or the quality of the planned trajectories. 

\subsubsection{Reference implementation and performance requirements}
The experiment uses the reference implementation (\secref{sec:implementation}). To provide a sufficient simulation update rate, the SDVPlanner instances target a planning rate of 3\,Hz, and the TrafficSim process targets updating the position of all vehicles at 30\,Hz.
Planning is a highly time-critical task, which needs to be executed within its target period of 333\,ms (3\,Hz). If a vehicle misses the target time to generate its plan, it likely affects the quality of its trajectory and the resulting motion. Furthermore, a long overrun can affect the SDV model's ability to predict the traffic state, resulting in sub-optimal trajectories and even unintended collisions. The state transformation and update is performed by TrafficSim and must be completed for all vehicles within 33\,ms (30\,Hz). A small exceedance, if consistent, may be acceptable, as it would slightly reduce the update frequency below 30\,Hz without compromising the actual vehicle motion. The experiment is executed on an Intel Core i7-6800K (3.40\,GHz), with 32\,GB RAM and Ubuntu 18.04.5.

\subsubsection{Scenarios}
We use two long-running scenarios, each with a two-minute duration, and vary the number of SDVs, up to 20. In each scenario, the SDVs travel in one lane and form a virtual platoon, simulating heavy traffic. In scenario A, the SDVs travel without any disturbance, and in scenario B, they need to steer to avoid a static obstacle in their lane. When running scenario B, the object collision checking for obstacle avoidance is activated. The purpose of scenario B is to show the impact of object collision checking on scalability, since it is computationally expensive. Each vehicle travelling behind another one is expected to observe a safe following distance.

\subsubsection{Metrics}
We evaluate the adherence to the target rates using the following metrics: \textit{Target Rate Compliance} (TRC), defined as the \% of simulation (execution) ticks from all vehicles that adhere to the target tick time of 33\,ms (30\,Hz); the \textit{maximum tick time}; the \textit{Target Planning Rate Compliance} (TPRC), defined as the \% of planning cycles from all vehicles that adhere to the target time of 333\,ms (3\,Hz); and the \textit{maximum planning time}.

\textit{\textbf{Results:}}\looseness=-1
Both scenarios with up to 20 vehicles execute successfully, without any collisions or lane boundary violations. The planning adheres to the target rate with almost 100\%, with 99.8\% being the worst case (Table \ref{tab:performance}). 
However, execution deteriorates significantly between 10 and 15 vehicles, especially when the collision checking is active, plunging from 98.49\% to 78.58\%.
Such a deterioration of the target rate to update the state of all vehicles may introduce inconsistencies and confuse the ADS under test, such as inducing significant errors in its object tracking system. However, reducing the update rate from 30\,Hz to 20\,Hz results in near perfect adherence for up to 20 vehicles when no collision checking is used and up to 15 vehicles with the collision checking active. Thus, scenarios with up to 10 SDV instances are easily handled by the reference implementation, and scaling to 20 instances requires reducing the update rate. For scenarios requiring more vehicles, the traffic can mix SDV instances for interactions with ego and vehicles based on predefined trajectories, with negligible computing cost.

\vspace{-2mm}
\begin{table}[t]
\centering
\caption{Performance with multiple scenario configurations}
\label{tab:performance}
\tabcolsep=0.05cm
\begin{tabular}{l|c c | c r r r r}
\toprule
id &vehicles &obstacle &coll. &TRC &max tick &TPRC &max plan\\
\midrule
3  &10 &inactive &0 &98.44\% &0.042s &100.00\% &0.333s\\
4  &15 &inactive &0 &92.28\% &0.055s &99.94\% &0.338s\\
5  &20 &inactive &0 &61.90\% &0.052s &100.00\% &0.333s\\
8  &10 &active &0 &98.49\% &0.041s &99.94\% &0.338s\\
9  &15 &active &0 &78.58\% &0.052s &99.80\% &0.340s\\
10 &20 &active &0 &55.65\% &0.065s &99.91\% &0.343s\\
\bottomrule
\end{tabular}
\vspace{-2mm}
\end{table}

\section{Conclusion}
\noindent 
\looseness=-1
We presented the SDV model to express and execute scenarios for ADS scenario-based testing in simulation. The model encapsulates driver and vehicle as a single entity with an architecture that provides a user-oriented language to coordinate the vehicle behavior and motion planning that optimizes for realism and achieving the scenario test objective. In particular, behavior trees provide a high-level description of discrete decisions, with a high-level of abstraction and parameterization to support controllability and reuse. Furthermore, dynamic trajectory planning allows for flexible adaptation of the SDV trajectories to different road geometries and achieving the test objective despite varying and unpredictable ego behaviors.

\looseness=-1
The evaluation shows that our proposed approach supports effective test scenario development and execution using the NHTSA vehicle-to-vehicle pre-crash scenarios, with high internal reuse of over 80\,\%. The analysis also shows that the majority of scenarios require dynamic trajectory planning, benefiting from the SDV model compared to the baseline.
The evaluation demonstrates the ability of the SDV model to faithfully reproduce real-world vehicle behavior, including different driving styles by adjusting parameters, with an average spatio-temporal trajectory distance of 1.24\,m. It also allows custom behaviors and misbehaviors by adding dedicated conditions and maneuvers. The reference implementation demonstrates that the SDV model scales to execute scenarios with 10-20 highly interactive vehicles, and additional optimizations, such as reducing the number of sampled trajectories for vehicles farther away from ego, allow for further scaling. Finally, the application of the SDV model to test WISE ADS in the cut-in scenario confirms the usefulness of the model and offers practical insights. Among others, the ability to control the shape of the cut-in trajectories uncovers the varied response of the ADS to different trajectories, showing that not only the target gap and velocity, but also the acceptance gap impact the likelihood of a collision. Furthermore, using an SDV model instance in place of ego helps accelerate the development of the test scenario and parameter selection to tune the trajectories of the agent that challenges ego.

\looseness=-1
In future work, we plan several model extensions and new capabilities that exploit the model. 
First, we plan to expand the model with new maneuvers and configuration options based on additional scenarios, harvested from a wider range of naturalistic data, such as the additional locations in the Waterloo dataset~\cite{online:wisetrafficdataset} and the multi-country INTERACTION dataset~\cite{ interactiondataset}. We plan to improve the auto-calibration process and further automate creation of behavior trees and their parameterization to approximate the naturalistic traffic. We will also expand the behavior trees and maneuvers for interaction with pedestrians~\cite{larter:2022:pedestrian}. Finally, we plan to exploit the model in generating new scenarios by injecting road-user misbehaviors into behavior trees, such as simulating distraction~\cite{VANLINT201863} and ignoring occlusions~\cite{kahn:2022:occlusion}. The SDV model implementation and toolset to design and run scenarios is publicly available and can be integrated with any simulation environment via co-simulation.

\section*{Acknowledgment}
\noindent\looseness=-1
The authors acknowledge the support of the Natural Sciences and Engineering Research Council of Canada, Renesas Electronics Corporation, Japan Science and Technology Agency ERATO Project ``HASUO Metamathematics for Systems, the PNRR MUR project VITALITY (ECS00000041), Spoke 2 ASTRA - ``Advanced Space Technologies and Research Alliance", of the PNRR MUR project CHANGES (PE0000020), Spoke 5 ``Science and Technologies for Sustainable Diagnostics of Cultural Heritage'', the PRIN project P2022RSW5W -
RoboChor: Robot Choreography, the PRIN project 2022JKA4SL - HALO: etHical-aware AdjustabLe autOnomous systems,
and of the MUR (Italy) Department of Excellence 2023 - 2027 for GSSI. The work of P.\ Pelliccione was also partially supported by the Centre of EXcellence on Connected, Geo-Localized and Cybersecure Vehicles (EX-Emerge), funded by the Italian Government under CIPE resolution n. 70/2017 (Aug. 7, 2017). This work was partially supported by the Wallenberg AI, Autonomous Systems and Software Program (WASP) funded by Knut and Alice Wallenberg Foundation.

\ifCLASSOPTIONcaptionsoff
  \newpage
\fi



\bibliographystyle{IEEEtran}
\bibliography{main_ieeejrnl}
%



%

\vskip -2\baselineskip 
\begin{IEEEbiography}[{\includegraphics[width=1in,height=1.25in,clip,keepaspectratio]{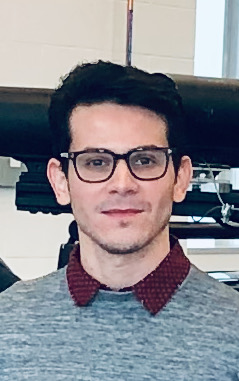}}]{Rodrigo Queiroz}
is a Ph.D. candidate at the University of Waterloo Faculty of Engineering (Canada) and is part of the Waterloo Intelligent Systems Engineering (WISE) Lab. His research interest focuses on validation \& verification of autonomous driving systems, safety-critical systems, motion planning, vehicle simulation, and traffic simulation. He received his Master's degree in Computer Science from the Federal University of Minas Gerais (UFMG) in Brazil.
\end{IEEEbiography}
\vskip -2\baselineskip 
\begin{IEEEbiography}[{\includegraphics[width=1in,height=1.25in,clip,keepaspectratio]{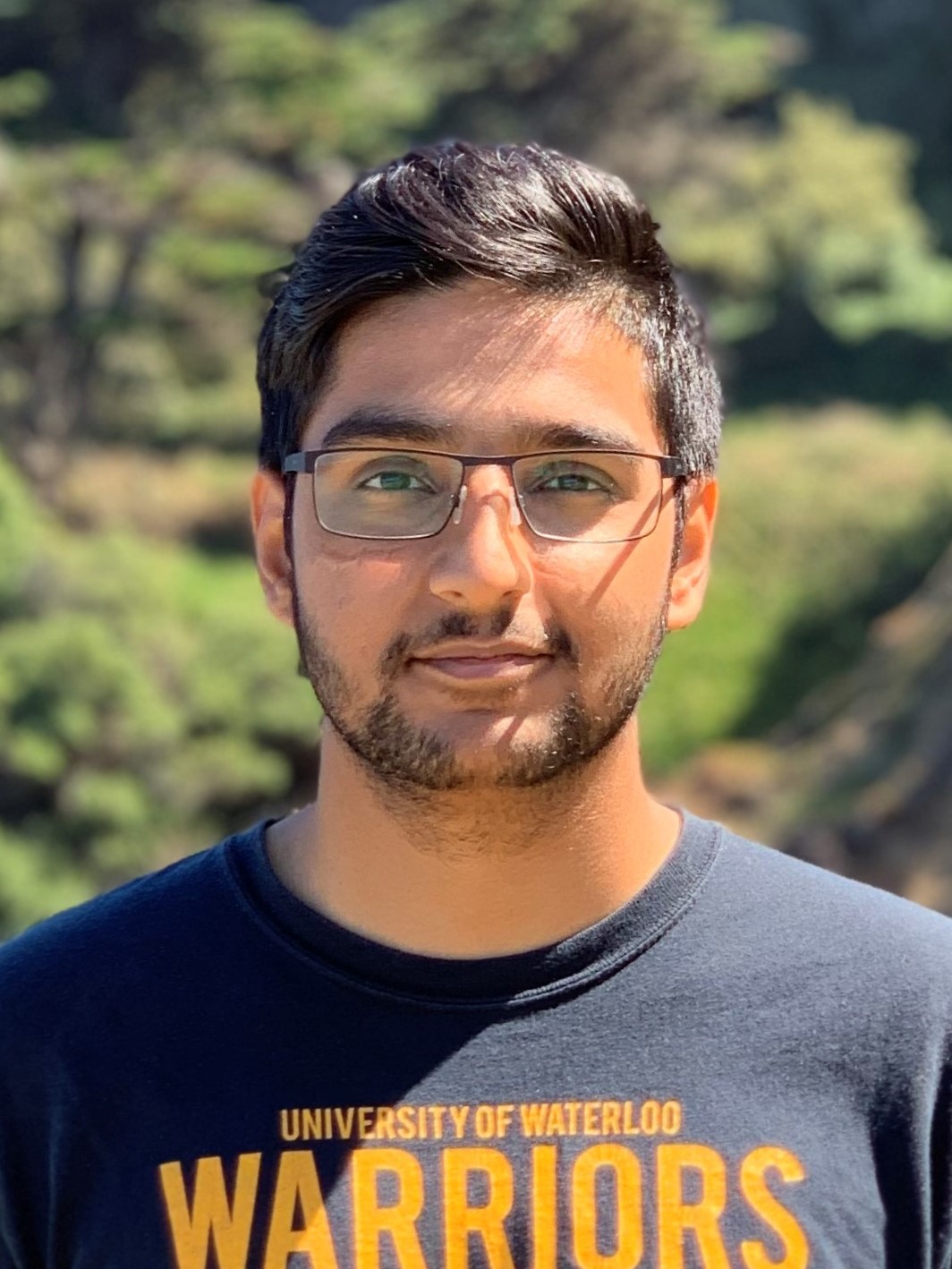}}]{Divit Sharma}
is a graduate from the University of Waterloo (Canada) with a Bachelor's degree in Computer Science. His interest and experience lie in robotics simulation and game development. His past work includes researching high-definition context maps for autonomous vehicles and developing interactive multi-agent simulation systems at the WISE Lab. He also has internship experience at Ike Robotics, a California-based self-driving startup, and Behaviour Interactive, a Montreal-based game development studio.
\end{IEEEbiography}
\vskip -1\baselineskip 
\begin{IEEEbiography}[{\includegraphics[width=1in,height=1.25in,clip,keepaspectratio]{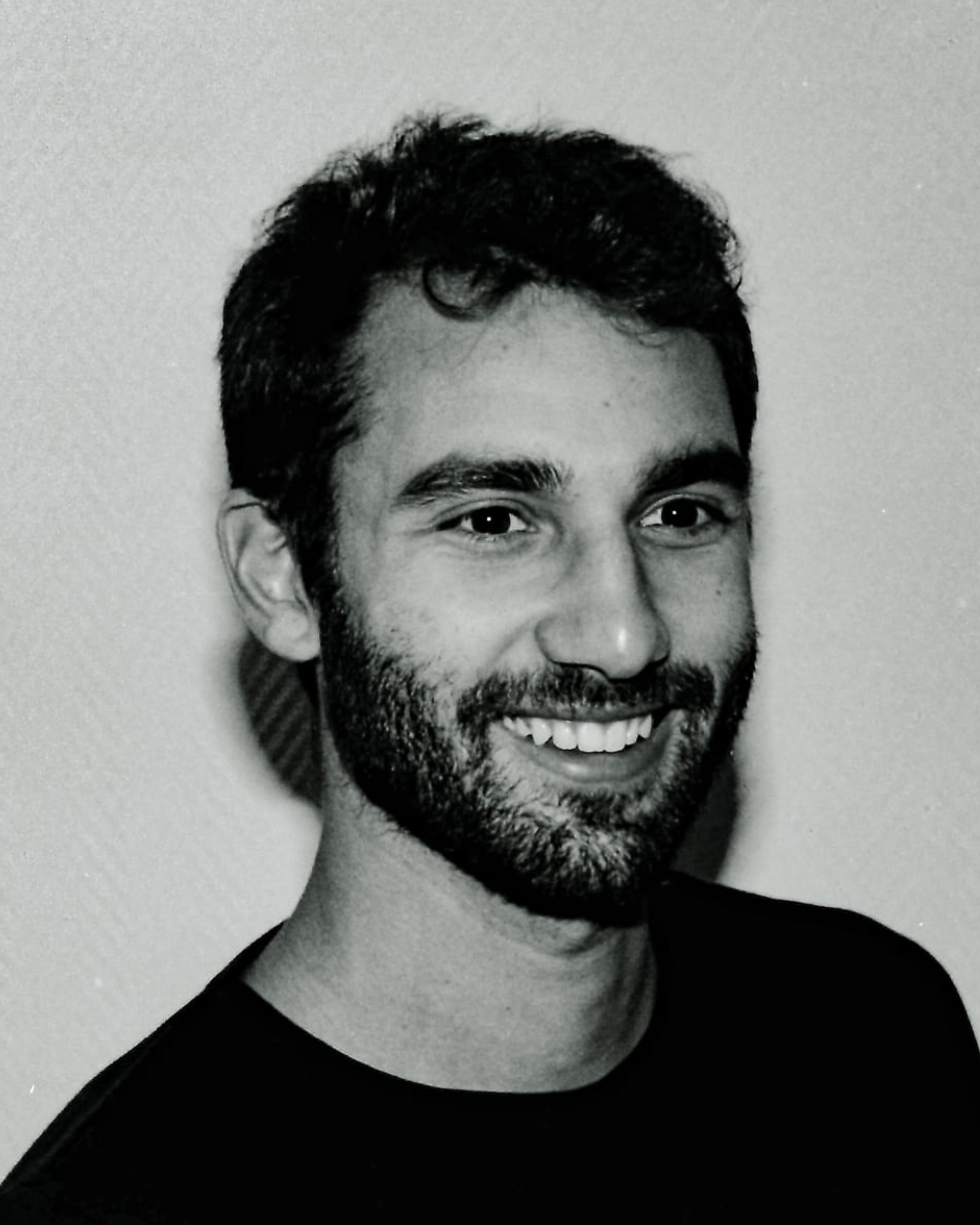}}]{Ricardo Caldas}
is a Ph.D. student at the Chalmers University of Technology in Sweden. 
His research interest focuses on verification of autonomous systems. Lately, he has been investigating the interplay between control theory and software engineering principles for the engineering of resilient autonomous systems, including mobile robots and self-driving vehicles. 
He received his master's degree in Computer Science from the University of Bras\'{i}lia in Brazil in 2019.
\end{IEEEbiography}
\vskip -2\baselineskip 
\begin{IEEEbiography}[{\includegraphics[width=1in,height=1.25in,clip,keepaspectratio]{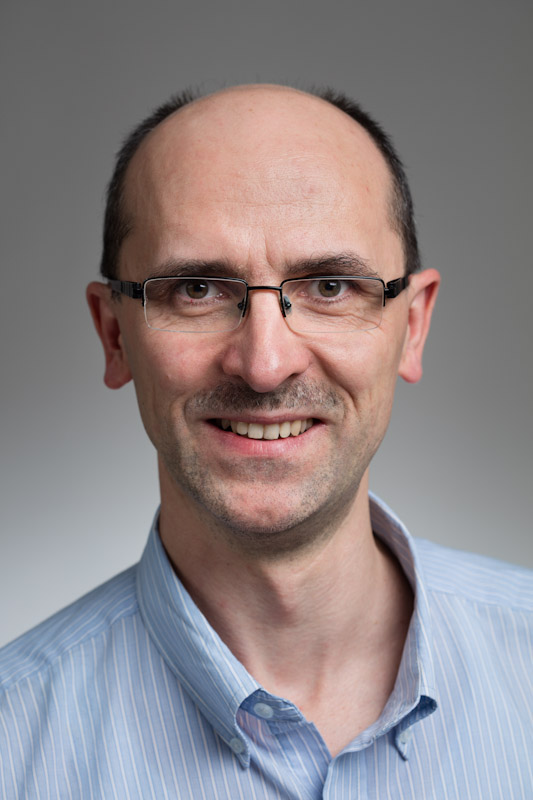}}]{Krzysztof Czarnecki} is a Professor of Electrical and Computer Engineering and a University Research Chair at the University of Waterloo, where he heads the Waterloo Intelligent Systems Engineering (WISE) Laboratory. He is a leading expert in the safety of automated driving systems (ADS), with focus on assuring the safety of driving behavior and machine-learned functions. He co-lead the development of the first ADS tested on public roads in Canada in 2018.
As a member of standardization committees, he has contributed to ISO 21448 (2nd edition), ISO 8800 (under development), and SAE J3164. He received the Premier's Research Excellence Award in 2004 and the British Computing Society in Upper Canada Award for Outstanding Contributions to IT Industry in 2008. He has also received eight Best Paper Awards, two ACM Distinguished Paper Awards, and four Most Influential Paper Awards.
\end{IEEEbiography}
\vskip -2\baselineskip 
\begin{IEEEbiography}[{\includegraphics[width=1in,height=1.25in,clip,keepaspectratio]{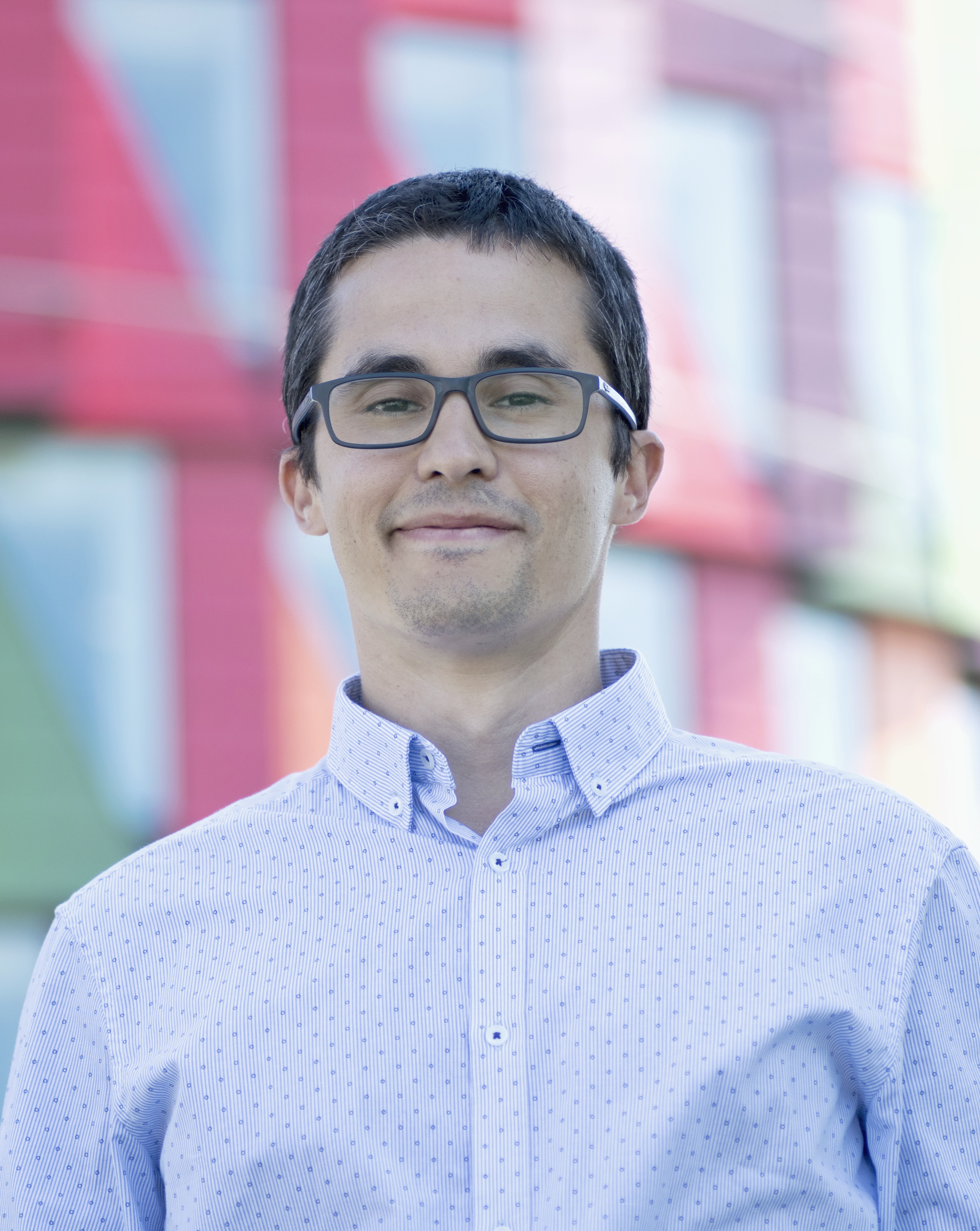}}]{Sergio Garc\'{i}a}
was a Ph.D. student at the University of Gothenburg in Sweden when working on this paper. He received the Ph.D. degree in September 2021.
His research interest focuses on robotics software engineering, striving to understand the complexity of the domain and analyzing its characteristics and challenges to develop solutions based on them.
He received his master's degree in electronics from the University of Alcalá in Spain in 2016.
\end{IEEEbiography}
\vskip -2\baselineskip 
\begin{IEEEbiography}[{\includegraphics[width=1in,height=1.25in,clip,keepaspectratio]{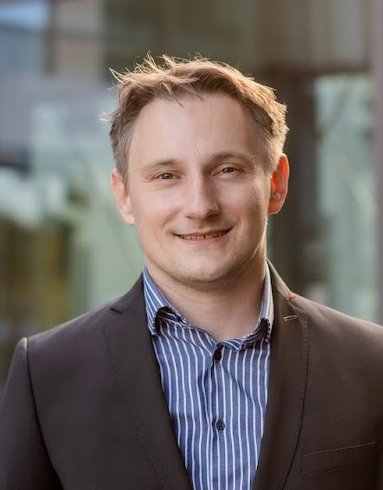}}]{Thorsten Berger}
 is a Professor in Computer Science at Ruhr University Bochum in Germany. His research focuses on automating software engineering for the next generation of intelligent, autonomous, and variant-rich software systems, exploring new ways of software creation, analysis, and evolution. He received the PhD degree in computer science from the University of Leipzig in Germany in 2013. Thereafter, he worked as a Postdoctoral Fellow at the University of Waterloo in Canada and the IT University of Copenhagen in Denmark, and as an Associate Professor at Chalmers\,$|$\,University of Gothenburg in Sweden. He received a fellowship from the Royal Swedish Academy of Sciences and the Wallenberg Foundation, one of the highest recognitions for researchers in Sweden. He received two best-paper and two most influential paper awards, as well as his service was recognized with distinguished reviewer awards at A* conferences.
\end{IEEEbiography}
\begin{IEEEbiography}[{\includegraphics[width=1in,height=1.25in,clip,keepaspectratio]{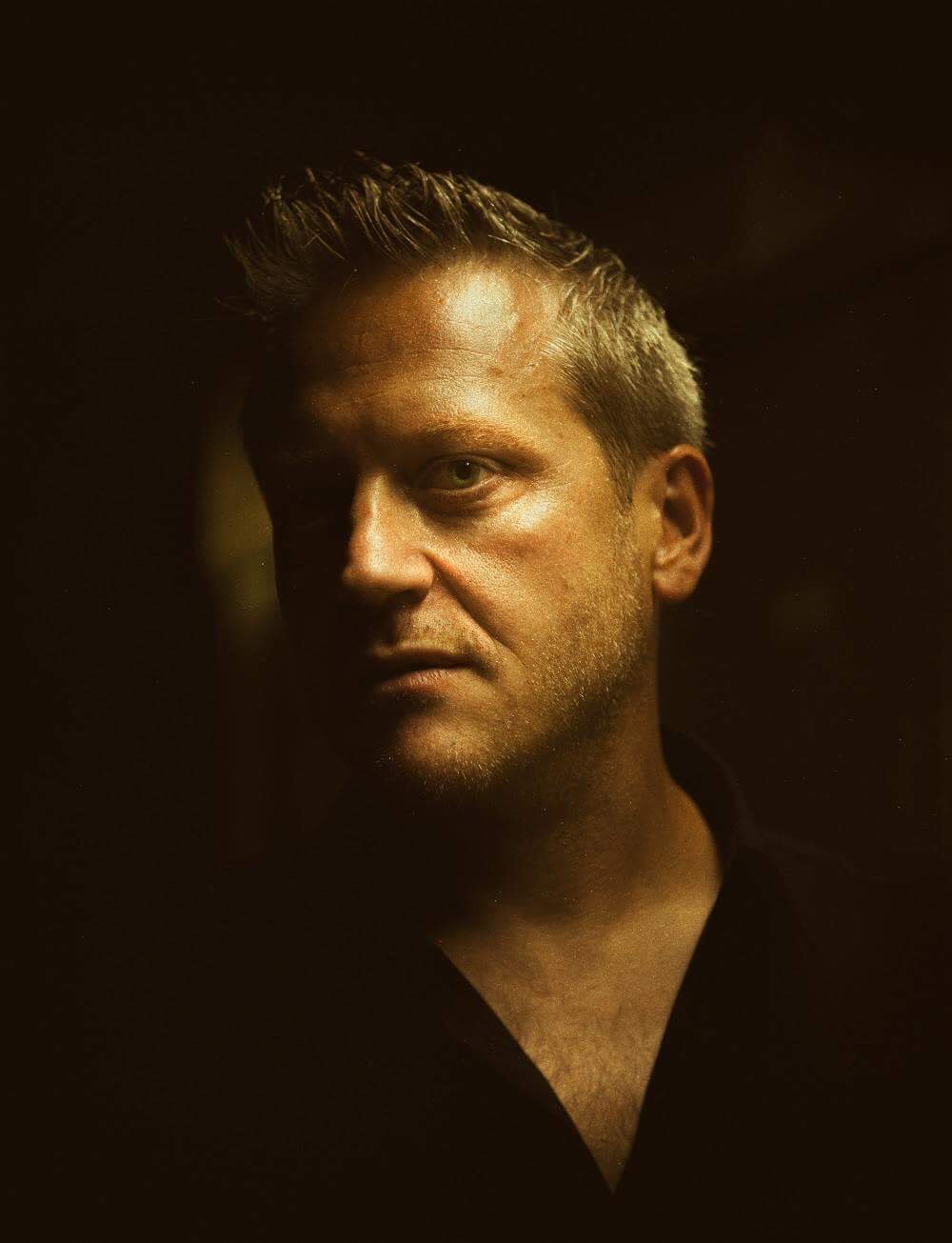}}]{Patrizio Pelliccione}
is a Professor in Computer Science and Director of the computer science area at Gran Sasso Science Institute (GSSI, Italy). He is also an adjunct professor at the University of Bergen, Norway. His research topics are mainly in software engineering, software architecture modeling and verification, autonomous systems, and formal methods. He received his Ph.D. in computer science from the University of L'Aquila (Italy). Thereafter, he worked as a senior researcher at the University of Luxembourg in Luxembourg, then assistant professor at the University of L'Aquila in Italy, then Associate Professor at both Chalmers $|$ University of Gothenburg in Sweden and University of L'Aquila.
He has been on the organization and program committees for several top conferences and he is a reviewer for top journals in the software engineering domain. He is very active in European and National projects. In his research activity, he has collaborated with several companies. More information is available at \url{http://www.patriziopelliccione.com}.

\end{IEEEbiography}





\vfill


\end{document}